\documentclass[letterpaper,twocolumn,10pt]{article}
\usepackage{pdfpages}
\usepackage{usenix-2020-09}
\usepackage{tikz}
\usepackage{amsmath}

\usepackage{filecontents}
\usepackage{hyperref}
\usepackage{makecell}
\usepackage{algorithm}
\usepackage{algorithmic}
\usepackage{wrapfig}
\usepackage{threeparttable}
\usepackage{graphicx}
\usepackage{bbding}
\usepackage{booktabs}
\usepackage{tablefootnote}
\usepackage{aisecure-math}
\usepackage{enumitem}
\usepackage{multirow}
\usepackage{url}
\definecolor{red}{rgb}{1,0,0}
\definecolor{darkgreen}{rgb}{0,0.5,0}
\definecolor{darkblue}{rgb}{0,0,0.5}
\definecolor{purple}{rgb}{1,0,1}
\newcount\Comments  
\newcommand{\kibitz}[2]{\ifnum\Comments=0\textcolor{#1}{#2}\fi}

\newcommand{\revision}[1]{\textcolor{black}{#1}}

\newcommand\underbf[1]{\underline{\textbf{#1}}}
\newenvironment{algocolor}{%
   \setlength{\parindent}{0pt}
   \color{black}
}{}
\usepackage{xspace}
\usepackage{enumitem}
\newcommand{\name}{\textsf{DiffSmooth}\xspace}
\newcommand{\namens}{\textsf{DiffSmooth}}

\begin{document}
\null%
\includepdf{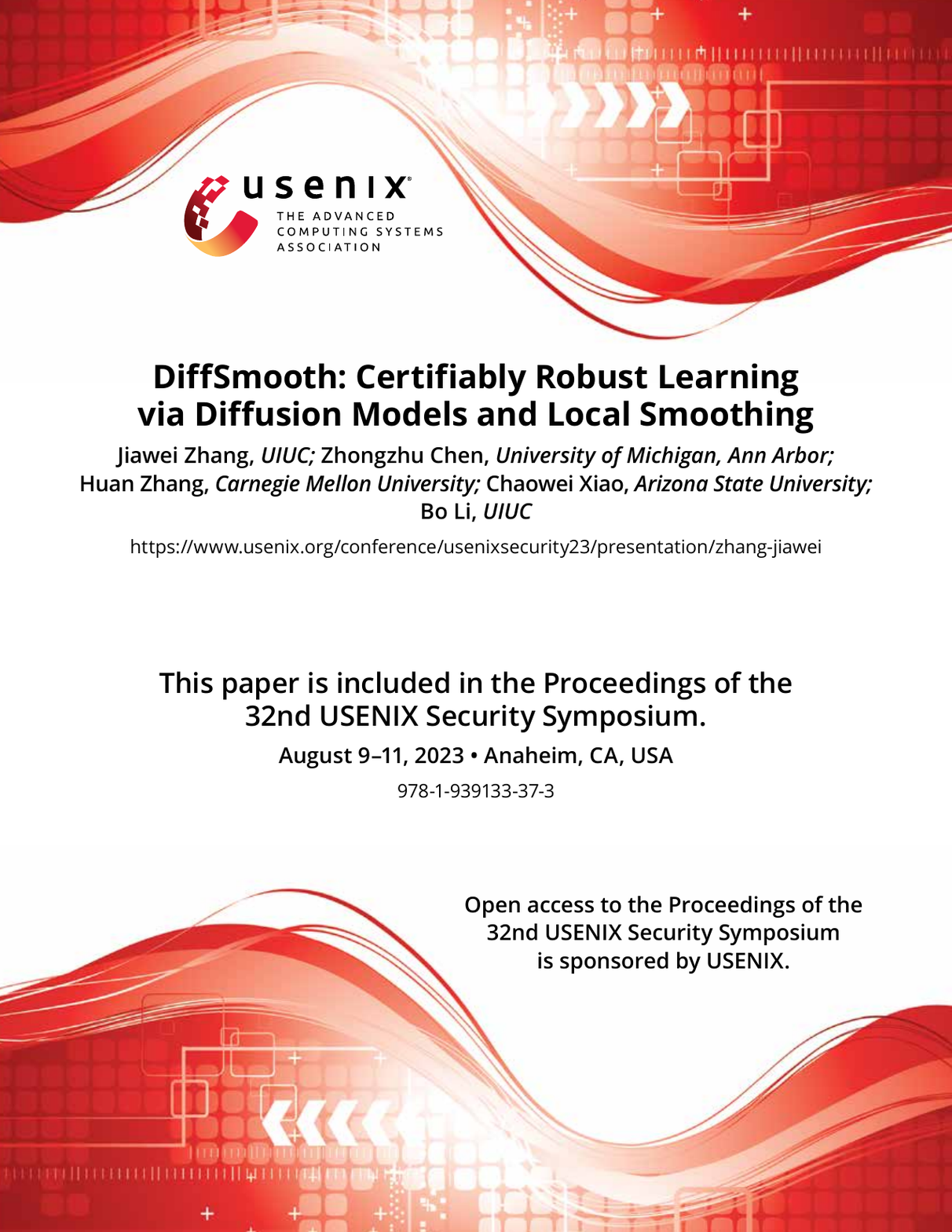}
\setcounter{page}{1}
\date{}

\title{\Large \bf \namens: Certifiably Robust Learning via Diffusion Models and Local Smoothing}

\author{
{\rm Jiawei Zhang}\\
UIUC
\and
{\rm Zhongzhu Chen}\\
University of Michigan, Ann Arbor
\and
{\rm Huan Zhang}\\
Carnegie Mellon University
\and
{\rm Chaowei Xiao}\\
Arizona State University 
\and
{\rm Bo Li}\\
UIUC
} 

\maketitle

\begin{abstract}
Diffusion models have been leveraged to perform adversarial purification and thus provide both empirical and certified robustness for a \textit{standard} model. On the other hand, different robustly trained \textit{smoothed} models have been studied to improve the certified robustness.
Thus, it raises a natural question: \textit{Can diffusion model be used to achieve improved certified robustness on those robustly trained smoothed models?} In this work, we first theoretically show that recovered instances by diffusion models are in the bounded neighborhood of the original instance with high probability; and the ``one-shot''  denoising diffusion probabilistic models (DDPM) can approximate the mean of the generated distribution of a continuous-time diffusion model, which approximates the original instance under mild conditions. Inspired by our analysis, we propose a certifiably robust pipeline \name, which first performs adversarial purification via diffusion models and then maps the purified instances to a common region via a simple yet effective \textit{local smoothing} strategy. 
We conduct extensive experiments on different datasets and show that \name  achieves SOTA-certified robustness compared with eight baselines. \revision{For instance, \name{} improves the SOTA-certified accuracy from $36.0\%$ to $53.0\%$ under $\ell_2$ radius $1.5$ on ImageNet.} The code is available at~\url{https://github.com/javyduck/DiffSmooth}.
\end{abstract}

\section{Introduction}

Despite the fact that the deep neural networks (DNNs) have achieved unprecedented success in different applications, they are still vulnerable to imperceptible adversarial noise, which will mislead the model to predict the perturbed input as an arbitrary adversarial target~\cite{biggio2013evasion,szegedy2014intriguing}. Such adversarial perturbations have posed a threat to the real-world application of DNNs on safety-critical scenarios such as fraud detection~\cite{pumsirirat2018credit,fiore2019using} and automatic driving~\cite{cao2021invisible,eykholt2018robust}.

Different \textit{empirical} defense approaches have been proposed to prevent such adversarial attacks. For instance, adversarial training~\cite{madry2018towards,wong2019fast,zhang2019theoretically,rebuffi2021fixing}, which incorporates adversarial instance into the training process, has become the de facto standard method for training robust models. However, empirical defenses may become broken under strong adaptive attacks~\cite{carlini2017adversarial,athalye2018obfuscated,croce2020reliable}.

Later, \textit{certified} defenses are proposed to provide a lower bound of accuracy for DNNs under constrained perturbations.
For instance, the technique of bound propagation~\cite{gowal2018effectiveness,zhang2019towards,wang2021beta}, which computes the upper and lower bounds of the features layer by layer, is commonly used to provide deterministic certification for small models and low-resolution instances. In addition, randomized smoothing~\cite{lecuyer2019certified,cohen2019certified,salman2019provably,zhai2019macer,jeong2020consistency} has been proposed as a more scalable technique for providing probabilistic certification on large-scale datasets, such as ImageNet~\cite{deng2009imagenet}, by taking a majority vote over the predictions of Gaussian-smoothed inputs.~\revision{Such technique typically requires to train a \textit{standard} model with Gaussian augmentation as a robustly trained  \textit{smoothed} model.}



At the same time, diffusion models~\cite{sohl2015deep,ho2020denoising} recently have demonstrated  powerful abilities of generative modeling in different tasks and applications, such as image generation~\cite{song2019generative,song2020improved,ho2020denoising}, shape generation~\cite{cai2020learning}, and image inpainting~\cite{song2020score}. In general, the diffusion model contains two processes: (1) the forward diffusion process, which perturbs the input data point with Gaussian noise gradually to populate low data density regions, and (2) the reverse diffusion process, which starts with random Gaussian noise and generates a high-quality instance through a Markov Chain iteratively. The denoising nature of the diffusion models has enabled a line of interesting works to purify adversarial perturbations and therefore improve the robustness of DNNs. For instance, Nie et al.~\cite{nie2022DiffPure}  proposed to purify  adversarial perturbations with diffusion models.

\revision{Given the promising diffusion-based adversarial purification, Lee~\cite{lee2021provable}~and Carlini et al.~\cite{carlini2022certified} propose a method Diffusion Denoised Smoothing (DDS), to leverage the denoising mechanism of diffusion models to remove the Gaussian noise added during the randomized smoothing process via a \textit{one-shot} reverse diffusion step. Thus, DDS is able to provide certified robustness for any off-the-shelf \textit{standard} model. Nevertheless, this approach can only provide state-of-the-art certified accuracy under small perturbation radii, and the robustness will decrease quickly for large perturbation radii (e.g., $r\ge 1.0$ in CIFAR-10 and $r\ge 2.0$ in ImageNet). Xiao et al.~\cite{xiao2022densepure} later attempt to use the \textit{multi-shot} reverse process and propose DensePure to repeat this process several times with different random seeds and take the majority vote over these purified images as the final prediction to further boost the certified robustness. However, as shown in~\Cref{sec:ablation}, even with a simple one-shot reverse diffusion step, the certification time for one image in ImageNet with sample size $N=10,000$ requires $553$s for purifying the Gaussian augmented images and only $11$s for prediction with a DNN model (i.e., ResNet-50~\cite{he2016deep}). Therefore, although DensePure performs much better than DDS, its actual computation cost is much higher than DDS owing to the repetitive multi-shot diffusion step.}

\revision{Based on existing observations, we aim to answer: \textit{(1) Can we improve the certified robustness of models under large perturbation radii leveraging the diffusion-based purification? (2) Unlike DensePure, can we further boost the certified robustness by executing more prediction steps instead of the reverse diffusion step, which is far more expensive?
}}

\revision{In this paper, we show that it is possible to achieve higher certified robustness with higher benign accuracy leveraging the robustly trained \textit{smoothed} model
based on our proposed local smoothing technique (formally introduced in Section~\ref{sec:method}).}

\revision{
In particular, we first provide theoretical analysis to show that the recovered instances from (adversarial) inputs will be in the bounded neighborhood of the corresponding original instance with high probability. 
We also prove that the ``one-shot" denoising of Denoising Diffusion Probabilistic Models (DDPM)~\cite{ho2020denoising} can approximate the mean of the generated posterior distribution by continuous-time diffusion models, which is, in turn, an approximation of the original instance under mild conditions.
}

\begin{figure*}[t]
    \centering
    \includegraphics[width=0.9\textwidth]{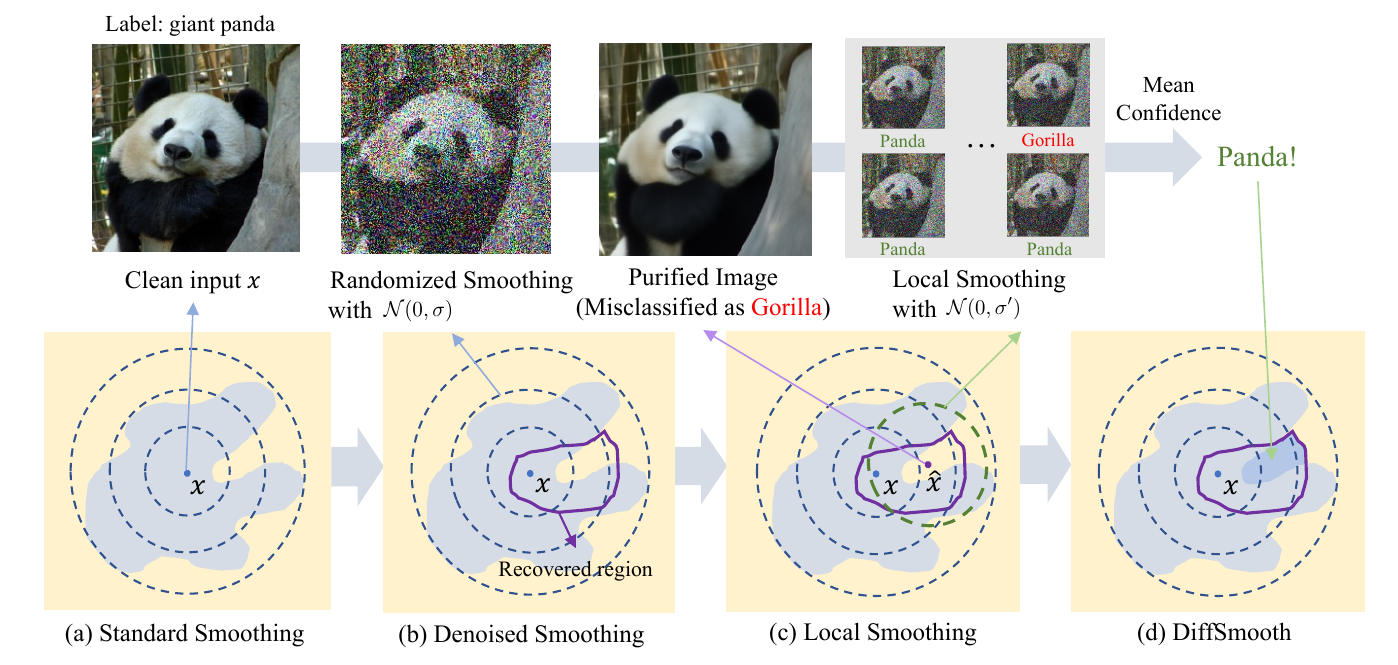}
    \vspace{-5mm}
    \caption{Overview of our pipeline~\name. The second row shows the decision regions of the base classifier for different smoothing processes at an input $x$. (a) follows the standard randomized smoothing~\cite{cohen2019certified}; (b) represents the denoised smoothing~\cite{salman2020denoised, carlini2022certified}, which first attempts to purify the noisy images with a denoiser, and then sends the purified image $\hat{x}$,~\revision{which is inside the purple recovered region}, to a \textit{standard} classifier for prediction. \name instead first performs local smoothing on the purified image $\hat{x}$ with a smaller noise level as shown by the ~\revision{green dash line} in (c), and then takes a majority vote over the local smoothed predictions as shown in (d).}
    \label{fig:pipeline}
    \vspace{-6mm}
\end{figure*}

\revision{
Inspired by our theoretical analysis of the properties of reversed instances and the relationship between the smoothed models and the robust regions of reversed instances, we propose a general certifiably robust adversarial purification pipeline \name.
In particular, as shown in~\Cref{fig:pipeline}, \name contains three steps: (1)  add a set of random Gaussian noise to the input $x$ for certification purposes; (2)  denoise each Gaussian perturbed input with the reverse process of a pre-trained diffusion model to generate a purified sample $\hat{x}$; (3) for each $\hat{x}$,  add another set of noise to generate locally smoothed instances and make predictions based on their mean confidence; (4)  repeat step (2)-(3) for all Gaussian perturbed inputs and take majority vote as the final \textit{smoothed} prediction.
An adversarial instance can be recovered to the neighborhood of the original one with high probability under mild conditions based on our analysis. Thus, adding a set of local smoothing noises to the recovered instance will help map it to a smoothed and robust region. }

Finally, we conducted comprehensive evaluations to compare the certified robustness of \name and seven SOTA baselines.
\revision{We specifically evaluate and control the  computation cost to compare with baselines in~\Cref{sec:ablation}}. We make the following technical contributions:
\begin{itemize}[leftmargin=*]
    \vspace{-1mm}
    \item We theoretically analyze the properties of purified adversarial instances of diffusion models. We prove that they are within the bounded neighborhood of the original clean instance with high probability, and their distances to the original instance depend on the adversarial perturbation magnitude and data density. We also prove that the ``one-shot'' denoising of DDPM can approximate the mean of the generated posterior distribution by continuous-time diffusion models, which is an approximation of the original instance under mild conditions.

    \vspace{-1mm}
    \item We show that naively combining diffusion models with smoothed models cannot effectively improve their certifiable robustness. Inspired by our theoretical analysis, we propose an effective and certifiably robust pipeline for smoothed classifiers, \name, via \emph{local smoothing}.


    \vspace{-1mm}
    \item We conduct extensive experimental evaluations on different datasets and show that \name achieves significantly higher certified robustness compared with SOTA baselines. For instance, \revision{with more inference steps}, the certified accuracy is improved from $36.0\%$ to $53.0\%$ under $\ell_2$ radius $1.5$ and $42.2\%$ to $48.2$ under $\ell_2$ radius $1.0$ on ImageNet; For CIFAR-10, the certified accuracy is improved from $42.8\%$ to $59.2\%$ under $\ell_2$ radius $0.50$, and from $39.4\%$ to $43.6\%$ under $\ell_2$ radius $1.00$. 

    \vspace{-1mm}
    \item We also perform a set of ablation studies to show that local smoothing is  unique to the diffusion purification process and evaluate the impacts of different parameters, such as the variance of local smoothing noise.~\revision{We show that \name outperforms DDS with the same computation cost.}
\end{itemize}

\vspace{-6mm}
\section{Related work}
\vspace{-2mm}

\paragraph{Certified robustness.} Deep neural networks (DNNs) are found vulnerable to adversarial examples~\cite{biggio2013evasion,szegedy2014intriguing}. To overcome such vulnerabilities, multiple empirical defenses have been proposed~\cite{papernot2016distillation,madry2018towards,samangouei2018defense,zhang2019theoretically}, most of which have been attacked again by strong adaptive attackers~\cite{carlini2017adversarial,athalye2018obfuscated,croce2020reliable}. 
Thus, certified robustness for DNNs is studied to provide a lower bound of model accuracy under constrained perturbations. While at the same time, the complete certification~\cite{zhang2018efficient,raghunathan2018certified, raghunathan2018semidefinite,salman2019convex,wang2021beta}, which guarantees to find the adversarial perturbation if it exists is constrained on a small dataset and extremely; and some incomplete certification~\cite{zhang2018efficient,salman2019convex,raghunathan2018certified,raghunathan2018semidefinite,gowal2018effectiveness,zhang2019towards} which may miss some certifiable instances are only applicable for specific model architectures and still can not scale to large datasets like ImageNet~\cite{deng2009imagenet}.
Later, Lecuyer et al.~\cite{lecuyer2019certified} prove robustness guarantee for smoothing with Gaussian and Laplace noise from a differential privacy perspective and first provide non-trivial certification result on ImageNet. The guarantee is later tightened by~\cite{cohen2019certified} as the robustness guarantee for randomized smoothing, which is \textit{probabilistic}. 
Based on this, the certification performance is further enhanced with the incorporation of adversarial training~\cite{salman2019provably}, consistency regularization~\cite{jeong2020consistency}, or ensemble model~\cite{liu2020enhancing,horvath2021boosting,yang2022on}. 


\vspace{-4mm}
\paragraph{Diffusion models for adversarial purification.} 
Diffusion models~\cite{sohl2015deep,ho2020denoising,song2020denoising,nichol2021improved} have shown impressive performance on generative modeling tasks and have been applied to various tasks such as image inpainting, super-resolution, and even text-to-image synthesis~\cite{dhariwal2021diffusion,saharia2022image,rombach2022high,bansal2022cold}.
The connection between the diffusion model and adversarial purification is first explored by~\cite{nie2022DiffPure} to remove adversarial perturbation, and~\cite{wang2022guided} further boosts the adversarial robustness with a guided reverse process. 
Besides, Carlini et al.~\cite{carlini2022certified} leveraged diffusion models to remove added smoothing Gaussian noise during randomized smoothing to provide certified robustness for a standard model,~\revision{and Xiao et al.~\cite{xiao2022densepure} propose to repeat the reverse diffusion steps multiple times and take the majority vote of the predictions on these reversed instances as the final prediction.}
\revision{One main limitation of existing work is that the certified robustness under large radii is low, given that the certification is calculated on standard models, which are less robust. In this work, we propose a novel local smoothing strategy to balance the tradeoff between low certified robustness on standard models and high computation cost on performing multi-reverse diffusion processes for a smoothed model. Our local smoothing strategy only requires multiple prediction steps, which are much cheaper than the reverse diffusion steps, while it helps to ``smooth'' the final predictions to improve robustness.
In addition,  we theoretically demonstrate that the reversed instances will lie in the vicinity of the original clean instance with a high probability. Finally, we show that our approach achieves state-of-the-art certified robustness compared with different baselines.
}



\vspace{-4mm}
\section{Background}

\vspace{-2mm}
\paragraph{Notations.} We mainly consider the classification problem in this paper. Let $\Delta^{k}$ be a $k$-dimensional probability simplex, we define the soft classifier $F:\mathbb{R}^d \rightarrow \Delta^{|\gY|}$ as the function which maps the input to a~\emph{confidence vector}. The associated hard classifier $f$ is defined as $f(x):=\argmax_{c\in \gY} F(x)$, which maps $\mathbb{R}^d$ to classes $\gY$.

\vspace{-4mm}
\paragraph{Robustness Certification.}
Given a radius $r \in \mathbb{R}_{+}$, the \textit{robustness certificate} provides a lower bound of the classification accuracy given perturbations within $r$, regardless of the concrete attack algorithms~\cite{liu2021algorithms, li2023sok}.
Formally, a certification algorithm takes a clean instance $x$ and the base  classifier $f$ as  inputs and  outputs a robust radius $r$, such that $f(x) = f(x')$ when the distance between $x$ and $x'$ satisfies $d(x,x') < r$, where $d(\cdot, \cdot)$ denotes a distance metric, e.g., the metric induced by $\ell_p$ norm. 
\revision{Generally, the implication of certified robustness is that it provides a lower bound of model robustness, given any perturbation whose magnitude is bounded by a $\ell_{p}$ norm. In other words, the empirical robustness under the same perturbation radius is always higher than the certified robustness, which is empirically tested in~\cite{hayes2020extensions}.}

\vspace{-4mm}
\setlength{\textfloatsep}{0pt}
\vspace{-6mm}
\begin{figure}[t]
\begin{algorithm}[H]
\caption{$\textsc{ComputeTimestep}(\sigma)$~\cite{carlini2022certified}.}
\label{alg:computet}
\begin{algorithmic}[1]
\renewcommand{\algorithmicrequire}{\textbf{Input:}}
 \renewcommand{\algorithmicensure}{\textbf{Output:}}
  \REQUIRE Magnitude of the smoothing noise $\sigma$.
  \ENSURE Start time step for the reverse process and the corresponding $\bar{\alpha}_t$.
    \STATE $t \gets 0$
    \WHILE{$\frac{1-\bar{\alpha}_t}{\bar{\alpha}_t}<\sigma^2$}
    \STATE $t\gets t+1$
    \ENDWHILE
  \STATE \textbf{return} $t, \bar{\alpha}_t$
 \end{algorithmic}
 \end{algorithm}
\end{figure}

\begin{figure}[t]
\vspace{-3em}
\begin{algorithm}[H]
\caption{$\textsc{Denoise}(x_t,t)$~\cite{carlini2022certified}. \# One-shot denoising}
\label{alg:denoise}
\begin{algorithmic}[1]
\renewcommand{\algorithmicrequire}{\textbf{Input:}}
 \renewcommand{\algorithmicensure}{\textbf{Output:}}
  \REQUIRE Intermediate sample $x_t$ and its associated timestep $t$.
  \ENSURE Predicted original clean image $\hat{x}_0$.
    \STATE $\epsilon \gets \epsilon_{\theta}(x_{t},t)$
    \STATE $\hat{x}_0 \gets \frac{1}{\sqrt{\bar{\alpha}_{t}}}(x_{t}-\sqrt{1-\bar{\alpha}_t}\epsilon)$
  \STATE \textbf{return} $\hat{x}_0$
 \end{algorithmic}
 \end{algorithm}
 \vspace{-1mm}
\end{figure}

\vspace{5mm}
\paragraph{Randomized Smoothing.}
In this paper, we mainly adopt \textit{randomized smoothing}~\cite{cohen2019certified} for achieving the certification of robustness. Specifically, randomized smoothing leverages the smoothed base classifier $f$ for prediction, which is defined by $g(x):=\argmax_{c\in \gY} \sP(f(x+\delta)=c)$ with $\delta \sim \mathcal{N}\left(\mathbf{0}, \sigma^{2} \mathbf{I}\right)$. Assume $\underline{p_A}$ is the lower bound of prediction probability of the top class $c_{A}$,
and $\overline{p_B}$ is the upper bound of prediction probability for the ``runner-up'' class,
then the smoothed classifier $g$ is robust around $x$ within the radius:
\vspace{-2mm}
\begin{equation}
 R=\frac{\sigma}{2}\left(\Phi^{-1}\left(\underline{p_{A}}\right)-\Phi^{-1}\left(\overline{p_B}\right)\right),
\label{eq:radius}
\vspace{-2mm}
\end{equation}
where $\Phi^{-1}$ is the inverse of the standard Gaussian CDF. Such certification requires the classifier to be robust under Gaussian noise, so usually, the base classifier will be trained with Gaussian augmentation~\cite{cohen2019certified}, and such robustly trained models are referred as \textit{smoothed models}.
Other robust training algorithms proposed later to further provide robustly trained smoothed models such as~\emph{SmoothAdv}~\cite{salman2019provably} and~\emph{Consistency}~\cite{jeong2020consistency}. 

\vspace{-3mm}
\paragraph{Denoised Smoothing.} To apply the off-the-shelf standard model with randomized smoothing, Salman et al.~\cite{salman2020denoised} propose to prepend a custom-trained denoiser $\mathcal{D}_\theta: \mathbb{R}^d \rightarrow \mathbb{R}^d$, which is trained to remove the Gaussian noise appeared in the instances, to the standard classifier $f_{\text{clf}}$, and treat $f_{\text{clf}} \circ \gD_\theta$ as the new base classifier $f$. In this way, the noisy input $x+\delta$ will first be purified by $\mathcal{D}_\theta$, and then the purified sample will be directly predicted with the pretrained standard classifier. As a result, it is expected that the prediction accuracy of  $f :=f_{\text{clf}} \circ \gD_\theta$ on the noisy instances with Gaussian perturbation is close to the accuracy of $f_{\text{clf}}$ given a clean instances when the denoiser $\gD_\theta$ performs well.

\setlength{\abovedisplayskip}{0pt}
\setlength{\belowdisplayskip}{0pt}
\setlength{\abovedisplayshortskip}{0pt}
\setlength{\belowdisplayshortskip}{0pt}

\vspace{-2mm}
\paragraph{Continuous-time Diffusion Models.}
A continuous-time diffusion model contains two components: (i) a diffusion process that adds random noises to the data gradually and finally researches a noise distribution, e.g., Gaussian distribution, and (2) a reverse process that removes the added noise to recover the original data distribution. The diffusion process can be defined by the stochastic differential equation:
\begin{align}\tag{SDE}
\label{SDE}
d \rvx = h(\rvx, t) dt+ g(t) d\rvw 
\end{align}
where $\rvx(0) \sim p$ (the original data distribution), $t\in [0,T]$, $h(\rvx, t)$ is the drift coefficient and $g(t)$ is the diffusion coefficient, and $\rvw(t)$ is the standard Wiener process \cite{anderson1982reverse}. Here we took the convention used by VP-SDE in \cite{song2020score} where $h(\rvx; t): =-\frac{1}{2} \gamma(t)\rvx$ and $g(t):= \sqrt{\gamma(t)}$ where $\gamma(t)$ is positive and continuous over $[0,T]$ such that 
$$\rvx(t) = \sqrt{\overline{\alpha}_t} \rvx(0)+ \sqrt{1-\overline{\alpha}_t} \boldsymbol{\epsilon}$$ 
where $\overline{\alpha}_t = e^{-\int_0^t \gamma(s)ds}$ and $\boldsymbol{\epsilon}\sim \mathcal{N}(\boldsymbol{0}, \boldsymbol{I})$. The reverse process can be defined by the stochastic differential equation: 
\begin{align}\tag{reverse-SDE}\label{reverse-SDE}
    d \hat \rvx = [h(\hat \rvx, t)- g(t)^2\nabla_{\hat \rvx}\log p_t(\hat \rvx) ] dt+ g(t) d\overline{\rvw}
\end{align}
where $dt$ denotes the infinitesimal reverse time step, and $\overline{\rvw}(t)$ is the reverse-time standard Wiener process. We use $\{\rvx(t)\}_{t\in [0,T]}$ and $\{\hat \rvx(t)\}_{t\in [0,T]}$ to denote the diffusion process and reverse process respectively. \cite{anderson1982reverse} shows that if $p\in \mathcal{C}^2$ and $\mathbb{E}_{\rvx\sim p} [||\rvx||_2^2]< \infty$, $\{\rvx(t)\}_{t\in [0,T]}$ and $\{\hat \rvx(t)\}_{t\in [0,T]}$ have the same distribution.

\vspace{-3mm}
\paragraph{Denoising Diffusion Probabilistic Models (DDPM)~\cite{ho2020denoising,nichol2021improved}} which construct the discrete forward diffusion process, has been shown effective to generate high-quality data through learning the reverse of the forward diffusion process. In particular, given the number of the forward steps $T$ and image $x_0$ sampled from the original data distribution, the forward diffusion acts to gradually adds a small amount of Gaussian noise following a variance schedule $\left\{\beta_t \right\}_{t=1}^T$, such that $q\left(\mathbf{x}_t | \mathbf{x}_{t-1}\right)=\mathcal{N}\left(\mathbf{x}_t ; \sqrt{1-\beta_t} \mathbf{x}_{t-1}, \beta_t \mathbf{I}\right)$. Generally,  $\beta_t$ is designed to increase with the time step $t$ and takes a value between $0$ and $1$; thus, the forward process will finally transform $x_0$ into an isotropic Gaussian noise. By leveraging the reparameterization trick, a  property of this forward process is that 
\begin{equation}
\label{eq:niceproperty}
    \rvx_t=\sqrt{\bar{\alpha}_t} \cdot x_0+\sqrt{1-\bar{\alpha}_t} \cdot \boldsymbol{\epsilon},
\end{equation}
where $\alpha_t:=1-\beta_t, \bar{\alpha}_t:=\prod_{i=1}^s \alpha_s$ and $\epsilon \sim \mathcal{N}(\mathbf{0}, \mathbf{I})$, which indicates that we can sample the intermediate noisy $x_t$ at any timestep $t$. Then, the diffusion model is trained to reverse the diffusion process and learn the posteriors  with the Markov chain $p_\theta\left(\mathbf{x}_{t-1}|\mathbf{x}_t\right)$, which is defined as $\mathcal{N}\left(\rvx_{t-1} ; \mu_\theta\left(x_t, t\right), \Sigma_\theta\left(x_t, t\right)\right)$. Here, $\mu_\theta\left(x_t, t\right) := \frac{1}{\sqrt{\alpha_t}}\left(x_t-\frac{\beta_t}{\sqrt{1-\bar{\alpha}_t}} \epsilon_\theta\left(x_t, t\right)\right)$ in which  $\epsilon_\theta$ is trained to predict the random noise $\epsilon$ for $x_t$, and  $\Sigma_\theta\left(x_t, t\right)$ is defined as $\sigma_t^2 \mathbf{I}$ as in~\cite{ho2020denoising} while it can also be learned following~\cite{nichol2021improved}. As a result, starting from $x_T\sim \mathcal{N}(\mathbf{0}, \mathbf{I})$, DDPM will generate an instance through the reverse sampling.

An interesting observation from~\cite{lee2021provable,carlini2022certified} is that this reverse process can be perfectly used to denoise Gaussian-corrupted images; hence it can be applied to the denoised smoothing~\cite{salman2020denoised} and acted as  $\gD_{\theta}$. Formally, given the corrupted instance $x_{rs}= x + \delta$ where $x$ is the clean instance and $\delta \sim  \mathcal{N}(\mathbf{0}, \sigma^2 \mathbf{I})$, we want to relate it to the noisy image $x_t$ sampled from the forward diffusion process with a specific timestep $t$, then we can hopefully remove the Gaussian noise $\delta$ to get the original instance $x$ with the reverse process which starts at $x_t$. To achieve this, notice that the intermediate sample from DDPM is shown in the form of $x_t=\sqrt{\bar{\alpha}_t} \cdot x+\sqrt{1-\bar{\alpha}_t} \cdot \epsilon$ where $\epsilon \sim  \mathcal{N}(\mathbf{0}, \mathbf{I})$; thus, if we scale $x_{rs}$ with $\sqrt{\bar{\alpha}_t}$ and equate the variance between the scaled $x_{rs}$ and $x_t$, we will obtain $\sigma^2 = \frac{1-\bar{\alpha}_t}{\bar{\alpha}_t}$. The solution of the timestep $t^*$ for this equation is straightforward: notice that the $\hat{\alpha}_t$ decreases monotonically with $t$ and $\hat{\alpha}_0 = 1$, and thus the value of $\frac{1-\bar{\alpha}_t}{\bar{\alpha}_t}$ will increase monotonically with $t$. Finally, the equation can be simply solved via 1D root-finding as shown in~\Cref{alg:computet}. As a result, we can simply start at $x_{t^*}=\sqrt{\bar{\alpha}_{t^*}}x_{rs}$ and perform the left reverse diffusion steps to recover the original $x$. In other words, we will recursively sample the previous intermediate image $\hat{x}_{t-1}$ based on the pre-defined Markov chain $p_\theta\left(\mathbf{x}_{t-1}|\mathbf{x}_t\right)$ until we get $\hat{x}_0$, which is exactly the purified image we want. However, the information of the $x$ contained in $x_{rs}$ will be destroyed in each iterative reverse diffusion process owing to the addition of the  Gaussian noise. In addition, Carlini et al.~\cite{carlini2022certified} propose to adopt one-shot denoising instead of running the full reverse diffusion process, where they first predict the likely $\epsilon$ in $x_{t^*}$ with $\epsilon_{\theta}$, and then directly plug it into~\Cref{eq:niceproperty} to obtain $\hat{x}_0$, for which the corresponding pseudo-code is shown in~\Cref{alg:denoise}.

\vspace{-2mm}
\section{\name: Diffusion Based Adversarial Purification with Local Smoothing}
\label{sec:method}
\vspace{-1mm}
In this section, we will~\revision{first provide the motivation of our method}, and then we analyze the diffusion-based adversarial purification in~\Cref{sec:theory}, which leverages \ref{reverse-SDE} to generate reversed instances, and we prove that such reversed instances will stay in the bounded neighborhood of the original clean instance with high probability. 
In addition to the reversed instance, to further understand the reversed posterior distribution, we will show that the ``one-shot'' denoising of DDPM (\Cref{alg:denoise}) will output the mean (approximation) of the conditional distribution generated by \ref{reverse-SDE}, based on the adversarial sample input and a given time step. 
We show that such a mean instance will have the ground-truth label of the corresponding original instance if the points with the ground-truth label have a high enough density in the original distribution.

Inspired by our theoretical analysis, we propose \name in~\Cref{sec:method}, consisting of diffusion-based adversarial purification and a simple yet effective local smoothing strategy. 

\vspace{-3mm}
\subsection{Motivation}
\label{sec:motivation}

\vspace{-1mm}
\revision{As shown in~\Cref{eq:radius}, the certified robustness largely depends on the \textit{consistency} of predictions of $N$ sampled points (larger $p_A$ will  result in larger certified radius $R$). Intuitively, the  “one-shot” reverse diffusion step in prior work DDS~\cite{carlini2022certified} helps to increase the prediction consistency by generating denoised samples near the original sample, for which we will provide the first formal theoretical analysis in~\Cref{sec:theory}. On the other hand, Xiao et al. \cite{xiao2022densepure} propose to repeat the reverse diffusion step multiple times and then take the majority vote of the predictions over these purified images as the final prediction. This leads to an approximation of consistent predictions for samples in the high-density region, thereby improving certified robustness.}

\revision{Nevertheless, as demonstrated in \Cref{sec:ablation}, the reverse diffusion step is actually the primary bottleneck for computation cost during certification. Thus, repeating the multi-shot reverse diffusion steps, as in \cite{xiao2022densepure}, will significantly increase the computation cost and render it impractical. Thus, we aim to propose a simple yet effective \textit{local smoothing} strategy by taking the majority vote of predictions on Gaussian smoothed samples given a smoothed model. This way, the computation cost for the reverse diffusion step is the same as DDS, while only extra computation cost for predictions is required, which is much lower. In the meantime, the consistency among the predictions will be improved since the distributions from which the Gaussian smoothed samples are drawn are sampled close.
In addition, given that a \textit{smoothed} model is more stable and therefore more robust than a \textit{standard} model at the cost of sacrificing benign accuracy, the local smoothing also helps to improve the benign accuracy of smoothed models.
}


\revision{Overall, \name first performs the one-shot diffusion-based adversarial purification, and then multiple  Gaussian noises are sampled to locally smooth the prediction for each purified sample. Such local smoothing will improve certified robustness for smoothed models and help to maintain or even improve benign accuracy since the smoothing Gaussian noises of the model and the locally smoothed samples are from similar distributions.}

\vspace{-4mm}
\subsection{Properties of Diffusion-Based Adversarial Purification}
\label{sec:theory}
\vspace{-2mm}
There are several works applying diffusion models to (adversarial) inputs by performing the diffusion and reverse processes on them, aiming to remove potential adversarial perturbations~\cite{carlini2017adversarial,nie2022DiffPure}. 
On the other hand, it is also possible to directly perform the reverse process to given inputs, and here we will analyze the properties and advantages of such reversed samples.
In particular, we theoretically prove that the reverse process of the diffusion model generates reversed samples in the bounded neighborhood of the original clean samples with high probability. We will analyze directly based on the stochastic equations \ref{SDE} and \ref{reverse-SDE}, as other diffusion models such as DDPM \cite{ho2020denoising} and score-based diffusion models \cite{song2020score} are approximations of the stochastic differential equations, and our results can also be extended to other models. 
Our main theorems are as follows.

\vspace{-3mm}
\begin{theorem}\label{reverseclose}
Given a data distribution $p \in \mathcal{C}^2$ and $\mathbb{E}_{\rvx\sim p} [||\rvx||_2^2]< \infty$. Let $p_t$ be the distribution of $\rvx(t)$ generated by \ref{SDE} and suppose $\nabla_{x} \log p_t(x) \le\frac{1}{2}C$ \revision{ for some constant $C$ and } $\forall t\in [0,T]$. Let $\gamma(t)$ be the coefficient defined in \ref{SDE} and $\overline{\alpha}_t = e^{-\int_0^t \gamma(s)ds}$. Then given an adversarial sample $x_{rs}= x_0 +\delta$ with original instance $x_0$ and perturbation $\delta$, solving \ref{reverse-SDE} starting at time $t^*$ and point $x_{t^*}=\sqrt{\bar{\alpha}_{t^*}}x_{rs}$ until time $0$ will generate a reversed random variable $\hat \rvx_0$ such that with a probability of at least $1-\eta$, we have
\begin{align}\label{thm1:ineq}
    ||\hat{\rvx}_0-x_0|| \leq ||x_{rs}-x_0||+\sqrt{e^{2 \tau\left(t^*\right)}-1} C_\eta+\tau\left(t^*\right) C
\end{align}
where $ \tau(t):= \int_0^t \frac{1}{2}\gamma(s) ds$, $C_{\eta}:= \sqrt{ d+2 \sqrt{d \log \frac{1}{\eta}}+2 \log \frac{1}{\eta}}$, and $d$ is the dimension of $x_0$.
\end{theorem}
\vspace{-6mm}
\begin{proof}(sketch)
Based on \cite[Theorem 3.2]{nie2022DiffPure}, we can obtain that
\begin{align*}
    ||\hat{\rvx}_0-x_0|| \leq ||\sqrt{\left(e^{ \tau\left(t^*\right)}-1\right)} \boldsymbol{\epsilon}+x_{rs}-x_0||+\tau\left(t^*\right) C
\end{align*}
where $\boldsymbol{\epsilon}\sim \mathcal{N}(\boldsymbol{0}, \boldsymbol{I})$. Since $||\boldsymbol{\epsilon}^2||\sim \chi^2(n)$, by the concentration inequality \cite{boucheron2013concentration}, we have
\begin{align*}
     \operatorname{Pr}\left(||\boldsymbol{\epsilon}|| \geq \sqrt{d+2 \sqrt{d \log \frac{1}{\eta}}+2 \log \frac{1}{\eta}}\right) \leq \eta.
\end{align*}
Thus, with probability at least $1-\eta$, we have 
\begin{align*}
   ||\hat{\rvx}_0-x_0|| \leq ||x_{rs}-x_0||+\sqrt{e^{2 \tau\left(t^*\right)}-1} C_\eta+\tau\left(t^*\right) C.
\end{align*}
\end{proof}
\vspace{-3mm}
For Theorem \ref{reverseclose} and \ref{oneshotbound}, please check Appendix \ref{theoproof} for complete proofs.

\noindent\textbf{Remark.}
Theorem \ref{reverseclose} implies that as long as $x_{rs}$ is not far away from the corresponding original instance $x_0$, $\overline{\alpha}_{t^*}$ is not very close to zero, and $\nabla_{x} \log p_t(x)$ is upper bounded by a reasonable value, \revision{the right-hand side of (\ref{thm1:ineq}) will be small. This means that} the reverse process of the diffusion model will generate a reversed sample in a small neighborhood of $x_0$ with high probability based on the scaled adversarial sample. \revision{Such examples are highly likely to have the same labels as $x_0$. One should note that since the diffusion model is a generative model with mode coverage on the whole dataset, it could be possible that $\hat{\rvx}_0$ lies far away from $x_0$. Theorem \ref{reverseclose}, developed based on a deep analysis of the stochastic differential equation \ref{reverse-SDE}, ensures that this unwanted case rarely happens.} Further, Theorem \ref{reverseclose} provides a tighter upper bound than \cite[Theorem 3.2]{nie2022DiffPure} in that $C_\eta$ is only half of the corresponding constant, which is due to the removal of the diffusion process (the reverse process is directly applied to a given input). 

In practice, we cannot implement the continuous-time diffusion model directly, and DDPM \cite{ho2020denoising} was proposed as one efficient approximation to the reverse process \ref{reverse-SDE}. In particular, DDPM learns a neural network $\epsilon_\theta(\rvx_t, t)$ to predict the likely noise added to $\rvx_0$ with the loss function:
\begin{align*}
    \mathbb{E}_{t, \rvx_0,  \boldsymbol{\epsilon}}\left[\frac{\beta_t^2}{2 \sigma_t^2 \alpha_t\left(1-\bar{\alpha}_t\right)}\left\| \boldsymbol{\epsilon}- {\epsilon}_\theta\left(\sqrt{\bar{\alpha}_t} \rvx_0+\sqrt{1-\bar{\alpha}_t}  \boldsymbol{\epsilon}, t\right)\right\|^2\right].
\end{align*}
where $t\in [0,T]$, $\rvx_0\sim p$, and $\boldsymbol{\epsilon}\sim \mathcal{N}(\boldsymbol{0}, \boldsymbol{I})$. For fast sampling in DDPM, the ``one-shot'' denoising (\cref{alg:denoise}) was frequently used \cite{ho2020denoising, carlini2022certified} where we have
\begin{align}\label{onestepDDPM}
    \hat{x}_0=\left(x_t-\sqrt{1-\bar{\alpha}_t}{\epsilon}_\theta\left(x_t, t\right)\right) / \sqrt{\bar{\alpha}_t}.
\end{align}
In the following theorem, we will show that given $t$ and $x_t$, the distance of $\hat x_0$ in (\ref{onestepDDPM}) to the mean of a conditional distribution generated by the reverse process \ref{reverse-SDE} starting at time $t$ and point $x_t$ will be bounded by the loss at time $t$:
\begin{align*}
    \ell_t(x_t):=\mathbb{E}_{\rvx_0, \boldsymbol{\epsilon}}\left[\frac{\beta_t^2}{2 \sigma_t^2 \alpha_t\left(1-\bar{\alpha}_t\right)}\left\| \boldsymbol{\epsilon}- {\epsilon}_\theta\left(x_t, t\right)\right\|^2\right.\\
    \left|\vphantom{\frac{\beta_t^2}{2 \sigma_t^2 \alpha_t\left(1-\bar{\alpha}_t\right)}\left\| {\epsilon}- {\epsilon}_\theta\left(x_t, t\right)\right\|^2}\sqrt{\bar{\alpha}_t} \rvx_0+\sqrt{1-\bar{\alpha}_t}  \boldsymbol{\epsilon} = x_t\right]
\end{align*}
where $\rvx_0\sim p$ and $\boldsymbol{\epsilon}\sim \mathcal{N}(\boldsymbol{0}, \boldsymbol{I})$.
\vspace{-2mm}
\begin{theorem}\label{oneshotbound}
Given a data distribution $p \in \mathcal{C}^2$ and $\mathbb{E}_{\rvx\sim p} [||\rvx||_2^2]< \infty$, given a time $t^*$ and point $x_{t^*}=\sqrt{\bar{\alpha}_{t^*}}x_{rs}$, the one-shot denoising for DDPM (\cref{alg:denoise}) will output an $\hat x_0$ such that
\begin{align}\label{thm2:ineq}
    \left\|\hat x_0 - \mathbb{E}\left[\hat \rvx_0\mid {\hat {\rvx}_{t^*} =  x_{t^*}}\right]\right \|\le \frac{2 \sigma_{t^*}^2 \alpha_{t^*}\left(1-\bar{\alpha}_{t^*}\right)^{3/2}}{\beta_{t^*}^2 \sqrt{\bar{\alpha}_{t^*}}}\cdot \ell_{t^*}(x_{t^*})
\end{align}
where $\hat \rvx_0, \hat \rvx_t$ are random variables generated by \ref{reverse-SDE}, $ \mathbb{P}\left(\hat{\rvx}_0 = x| {\hat {\rvx}_t = x_{t^*}}\right) \propto p(x) \cdot  \frac{1}{\sqrt{\left(2\pi\sigma^2_t\right)^n}} \exp\left({\frac{-|| x -x_{t^*}||^2_2}{2\sigma^2_t}}\right)$ and $\sigma_t^2 = \frac{1-\alpha_t}{\alpha_t}$ is the variance of Gaussian noise added at time $t$ in the diffusion process.
\end{theorem}
\vspace{-4mm}
\begin{proof}(sketch) Under the assumptions that $p \in \mathcal{C}^2$ and $\mathbb{E}_{\rvx\sim p} [||\rvx||_2^2]< \infty$, the diffusion process by \ref{SDE} and the reverse process by \ref{reverse-SDE} follow the same distribution ideally. Therefore, $\mathbb{P}\left(\hat{\rvx}_0 = x| {\hat {\rvx}_t = x_{t^*}}\right) =\mathbb{P}\left({\rvx}_0 = x| {{\rvx}_t = x_{t^*}}\right) \propto p(x) \cdot  \frac{1}{\sqrt{\left(2\pi\sigma^2_t\right)^n}} \exp\left({\frac{-|| x -x_{t^*}||^2_2}{2\sigma^2_t}}\right)$. Since $\sqrt{\bar{\alpha}_{t^*}} \rvx_0+\sqrt{1-\bar{\alpha}_{t^*}}  \boldsymbol{\epsilon} = x_{t^*}$ implies $\sqrt{\bar{\alpha}_{t^*}} \hat \rvx_0+\sqrt{1-\bar{\alpha}_{t^*}}  \boldsymbol{\epsilon} = x_{t^*}$ where $\boldsymbol{\epsilon} \sim \mathcal{N}(\boldsymbol{0}, \boldsymbol{I})$, we have
\begin{align*}
     \ell_{t^*}(x_{t^*})=&~\mathbb{E}_{\rvx_0, \boldsymbol{\epsilon}}\left[\frac{\beta_{t^*}^2 \sqrt{\bar{\alpha}_{t^*}}}{2 \sigma_{t^*}^2 \alpha_{t^*}\left(1-\bar{\alpha}_{t^*}\right)^{3/2}}\left\| \hat \rvx_0 - \hat x_0\right\|\right|\left.\vphantom{\frac{\beta_{t^*}^2 \bar{\alpha}_{t^*}}{2 \sigma_{t^*}^2 \alpha_{t^*}\left(1-\bar{\alpha}_{t^*}\right)^{3/2}}}\rvx_{t^*} = x_{t^*}\right]\\
     \ge &~ \frac{\beta_{t^*}^2 \sqrt{\bar{\alpha}_{t^*}}}{2 \sigma_{t^*}^2 \alpha_{t^*}\left(1-\bar{\alpha}_{t^*}\right)^{3/2}}\cdot  \left\|\hat x_0 - \mathbb{E}\left[\hat \rvx_0\mid {\hat {\rvx}_{t^*} =  x_{t^*}}\right]\right \|,
\end{align*}
where the last inequality is by Jensen's inequality \cite{boyd2004convex}.
\end{proof}
\vspace{-1mm}
 \noindent \textbf{Remark.} 
\revision{The right-hand side of (\ref{thm2:ineq}) is the multiplication of a constant depending on $t^*$ and the loss $\ell_{t^*}(x_{t^*})$ at time $t^*$. This implies that if we have smaller (zero) loss $\ell_{t^*}(x_{t^*})$ at time $t^*$,} \revision{$\hat x_0$ will approximate the mean of the conditional distribution $ \mathbb{P}\left(\hat{\rvx}_0 = x| {\hat {\rvx}_t = x_{t^*}}\right)$.} In addition, this conditional distribution has a high density on points with high data density in the original distribution $p$ and close to $x_{rs}$. Such points tend to have the same ground-truth label as the original instance and can be recognized well by the classifiers trained on the data manifold. That means, as long as the original clean instance lies in a high enough data density region in the original distribution, the mean of the generated conditional distribution will have a similar property (i.e., prediction label) as such a clean instance. 



\vspace{-3mm}
\subsection{Certifying Smoothed Models with \name}
\label{sec:procedure}
\vspace{-1mm}
Based on our theoretical analysis above, it is clear that the reversed samples are in the bounded neighborhood of the corresponding clean instance. Thus, given a robustly trained \textit{smoothed} model, in order to further improve its certified robustness by improving its clean accuracy,  we propose a simple yet effective \textit{local smoothing} technique for the reversed samples based on diffusion models. In this section, we will describe in detail how~\name works. 
 \begin{algorithm}[t]
\caption{$\textsc{PurifyClassifier}(x_{rs};\sigma',m)$.}
\label{alg:purifyf}
\begin{algorithmic}[1]
\renewcommand{\algorithmicrequire}{\textbf{Input:}}
 \renewcommand{\algorithmicensure}{\textbf{Output:}}
  \REQUIRE The input noisy sample $x_{rs}$, the magnitude of the local smoothing noise $\sigma'$, the number of sampled local smoothing noise $m$.
  \ENSURE The prediction for $x_{rs}$.
  \STATE $t^*,\bar{\alpha}_{t^*} \gets \textsc{ComputeTimestep}(2\sigma)$
  \STATE $\hat{x} \gets \left(\textsc{Denoise}(\sqrt{\bar{\alpha}_{t^*}}(2x_{rs}-1),t^*)+1\right)/2$
  \STATE $\hat{y} \gets 0$
\FOR{$i = 1$ to $m$}
\STATE $\delta_i' \sim \gN \left(0, \sigma'^2 \mathbf{I}\right)$
\STATE $\hat{y} \gets \hat{y} + \frac1m F(\hat{x}+\delta_i')$
\ENDFOR
  \STATE \textbf{return} $\argmax_{c\in \gY} \hat{y}$
 \end{algorithmic}
 \end{algorithm}

\begin{algorithm}[t]
\caption{$\textsc{SampleUnderNoise}(f, x, n, \sigma)$~\cite{cohen2019certified}.}
\label{alg:sample}
\begin{algorithmic}[1]
\renewcommand{\algorithmicrequire}{\textbf{Input:}}
 \renewcommand{\algorithmicensure}{\textbf{Output:}}
  \REQUIRE Base classifier $f$, clean input image $x$, the number of smoothing noise $n$, smoothing noise magnitude $\sigma$.
  \ENSURE A vector of class counts.
\STATE $\texttt{counts} \gets [0,0,...,0]$
\FOR{$i = 1$ to $n$}
\STATE $x_{rs} \gets x + \mathcal{N}\left(\mathbf{0}, \sigma^{2} \mathbf{I}\right)$
\STATE $y \gets f(x_{rs})$
\STATE $\texttt{counts}[y] += 1$
\ENDFOR
  \STATE \textbf{return} \texttt{counts}
 \end{algorithmic}
 \end{algorithm}
 
\begin{algorithm}[t]
   \caption{Certification Process for~\name.}
   \begin{algorithmic}[1]
\renewcommand{\algorithmicrequire}{\textbf{Input:}}
 \renewcommand{\algorithmicensure}{\textbf{Output:}}
  \REQUIRE The magnitude of the smoothing noise $\sigma$, the magnitude of the local smoothing noise $\sigma'$, the number of sampled local smoothing noise $m$, the number of the smoothing noise for selection $n_0$, the number of the smoothing noise for estimation $n$, the certification confidence $(1-\alpha)$. 
  \ENSURE Certified prediction and its robust radius.
   \label{alg:certification}
   \STATE $f \gets \textsc{PurifyClassifier}(\ \cdot\ ; \sigma',m)$.
   \STATE $\texttt{counts0} \leftarrow \textsc{SampleUnderNoise}(f, x, n_0, \sigma)$
   \STATE $\hat{c}_A \leftarrow$ top index in \texttt{counts0}
   \STATE $\texttt{counts} \leftarrow \textsc{SampleUnderNoise}(f, x, n, \sigma)$
   \STATE $\underline{p_A} \leftarrow \textsc{LowerConfBound}$($\texttt{counts}[\hat{c}_A]$, $n$, $1 - \alpha$) 
   \IF{ $\underline{p_A} > \frac{1}{2}$}
   \STATE \textbf{return} $\hat{c}_A$ and radius $\sigma \, \Phi^{-1}(\underline{p_A})$ 
   \ELSE 
   \STATE \textbf{return} ABSTAIN
   \ENDIF
\end{algorithmic}
\end{algorithm}

\renewcommand\arraystretch{1.15}
\begin{table*}[t]
\centering
\caption{\small Certified accuracy of ResNet-110 on CIFAR-10 under different $\ell_2$ radii. The smoothed model used for our method~\name is indicated inside the brackets, e.g.,~\namens(Gaussian) indicates the base smoothed model is trained with~\emph{Gaussian}.}
\resizebox{0.85\linewidth}{!}{
\begin{threeparttable}
\begin{tabular}{cccccccccccc}
\toprule\hline
\multirow{2}{*}{Method\tnote{1}}& \multirow{2}{*}{Extra data} & \multicolumn{10}{c}{Certified Accuracy (\%) under $\ell_2$ Radius $r$}   \\
& & 0.00 & 0.25 & 0.50 & 0.75 & 1.00 & 1.25 & 1.50 & 1.75 & 2.00 & 2.25 \\ \hline
\multicolumn{1}{c|}{Gaussian~\cite{cohen2019certified}} & \tiny\XSolidBrush& 75.0 & 60.0 & 42.8 & 32.0 & 23.0 & 17.4 & 14.0 & 11.8 & 9.8 & 7.6 \\
\multicolumn{1}{c|}{SmoothAdv~\cite{salman2019provably}} & \tiny\XSolidBrush& 73.6 & 66.8 & 57.2 & 47.2 & 37.6 & 32.8 & 28.8 & 23.6 & 19.4 & 16.8 \\
\multicolumn{1}{c|}{SmoothAdv~\cite{salman2019provably}} & \checkmark& 80.8 & 71.4 & 63.2 & 52.6 & 39.4 & 32.2 & 26.2 & 22.2 & 20.2 & 18.4 \\
\multicolumn{1}{c|}{MACER~\cite{zhai2019macer}} & \tiny\XSolidBrush& 81.0 & 71.0 & 59.0 & 47.0 & 38.8 & 33.0 & 29.0 & 23.0 & 19.0 & 17.0 \\
\multicolumn{1}{c|}{Consistency~\cite{jeong2020consistency}} & \tiny\XSolidBrush& 77.8 & 68.8 & 58.1 &48.5 & 37.8 & 33.9 & 29.9 &  25.2 & 19.5 & 17.3 \\
\multicolumn{1}{c|}{SmoothMix~\cite{jeong2021smoothmix}} &\tiny\XSolidBrush & 77.1 & 67.9 & 57.9 & 47.7 & 37.2 & 31.7 & 25.7 & 20.2 & 17.2 & 14.7 \\
\multicolumn{1}{c|}{Boosting~\cite{horvath2021boosting}}  &\tiny\XSolidBrush & 83.4 & 70.6 & 60.4 & 52.4 & 38.8 & 34.4 & 30.4 & 25.0 & 19.8 & 16.6 \\
\multicolumn{1}{c|}{DDS(Standard)~\cite{carlini2022certified}\tnote{2}}  &\tiny\XSolidBrush & 79.0 & 62.0 & 45.8 & 32.6 & 25.0 & 17.6 & 11.0 & 6.2 & 4.2 & 2.2 \\
\multicolumn{1}{c|}{DDS(Smoothed)~\cite{carlini2022certified}\tnote{3}}  &\checkmark & 79.8 & 69.9 & 55.0 & 47.6 & 37.4 & 32.4 & 28.6 & 24.8 & 15.4 & 13.6 \\ 
\multicolumn{1}{c|}{\namens(Gaussian)}  &\tiny\XSolidBrush & 78.2 & 67.2 & 59.2 & 47.0 & 37.4 & 31.0 & 25.0 & 19.0 & 16.4 & 14.2 \\
\multicolumn{1}{c|}{\namens(SmoothAdv)} &\tiny\XSolidBrush & 82.8 & 72.0 & 62.8 & 51.2 & 41.2 & 36.2 & \underline{\textbf{32.0}} & \underline{\textbf{27.0}} & \underline{\textbf{22.0}} & 19.0 \\
\multicolumn{1}{c|}{\namens(SmoothAdv)} & \checkmark& \underline{\textbf{85.4}} & \underline{\textbf{76.2}} & \underline{\textbf{65.6}} & \underline{\textbf{57.0}} & \underline{\textbf{43.6}} & \underline{\textbf{37.2}} & 31.4 & 25.2 & 21.6 & \underline{\textbf{20.0}} \\ \hline\bottomrule
\end{tabular}
\begin{tablenotes}
\footnotesize
\item[1] We report the performance for Gaussian and SmoothAdv based on pretrained models.
\item[2] We reimplement and report the results of DDS~\cite{dosovitskiy2020image} on ResNet-110.
\item[3] We use the same smoothed models as tested on \name (i.e., Gaussian and SmoothAdv) for DDS and report the best results.
\end{tablenotes}
\end{threeparttable}}
\label{tab:cifar}
\vspace{-6mm}
\end{table*}

\renewcommand\arraystretch{1.15}
\begin{table*}[t]
\centering
\caption{\small Certified accuracy on ImageNet under different $\ell_2$ radii. The smoothed model used for our method~\name is indicated inside the brackets, e.g.,~\namens(Gaussian) indicates the base smoothed model is trained with~\emph{Gaussian}.}
\begin{threeparttable}
\renewcommand{\TPTminimum}{\linewidth}
\makebox[\linewidth]{
\resizebox{0.75\linewidth}{!}{
\begin{tabular}{c|c|ccccccc}
\toprule\hline
\multirow{2}{*}{Architecture} & \multirow{2}{*}{Method\tnote{1}}         & \multicolumn{7}{c}{Certified Accuracy (\%) under $\ell_2$ Radius $r$} \\
& & 0.00 & 0.50 & 1.00 & 1.50 & 2.00 & 2.50 & 3.00 \\ \hline
\multirow{10}{*}{ResNet-50} & Gaussian~\cite{cohen2019certified} & 66.4 & 48.6 &37.0 &25.4 &18.4 &13.8 &10.4 \\
& SmoothAdv~\cite{salman2019provably} & 66.6 & 52.6 & 42.2 & 34.6 & 25.2 & 21.4 & 18.8 \\
& MACER~\cite{zhai2019macer} & \underline{\textbf{68.0}} & 57.0 & 43.0 & 37.0 & 27.0 & 25.0 & 20.0 \\
& Consistency~\cite{jeong2020consistency} & 57.0 & 50.0 & 44.0 & 34.0 & 24.0 & 21.0 & 17.0 \\
& SmoothMix~\cite{jeong2021smoothmix}   & 55.0 & 50.0  &43.0 & 38.0 & 26.0 & 24.0 & 20.0 \\
& Boosting~\cite{horvath2021boosting}\tnote{2}  & \underline{\textbf{68.0}} & 57.0 & 44.6 & 38.4 & 28.6 & 24.6 & 21.2 \\
& DDS(Standard)~\cite{carlini2022certified}\tnote{3}    &67.4 &49.0 & 33.0 & 22.2 &17.4 &12.8 &8.0 \\ 
& DDS(Smoothed)~\cite{carlini2022certified}\tnote{4}   & 48.0 & 40.6 & 29.6 & 23.8 & 18.6 & 16.0 & 13.4 \\ \cline{2-9}
& \namens(Gaussian)  &66.2 & 57.8 &44.2 &36.8 & 28.6&25.0 &19.8 \\
& \namens(SmoothAdv) & 66.2 & \underline{\textbf{59.2}} & \underline{\textbf{48.2}} & \underline{\textbf{39.6}} & \underline{\textbf{31.0}} & \underline{\textbf{25.4}} & \underline{\textbf{22.4}} \\ \hline\hline
\multirow{4}{*}{\makecell{BEiT\tnote{6}}} & Gaussian~\cite{cohen2019certified} & 82.0 & 70.2 & 51.8 & 38.4 & \revision{32.0} & \revision{23.0} & \revision{17.0} \\
& DDS(Standard)~\cite{carlini2022certified}    & 82.8 &71.1 & 54.3  &  38.1&   \revision{29.5} &  \revision{-} &   \revision{13.1}\\ 
& DDS(Smoothed)~\cite{carlini2022certified}    & 76.2 & 60.2  & 43.8 & 31.8  &  \revision{22.0} &  \revision{17.8}&  \revision{12.2} \\ 
\cline{2-9}
& \namens(Gaussian)    & \underline{\textbf{83.8}} & \underline{\textbf{77.2}} & \underline{\textbf{63.2}} & \underline{\textbf{53.0}}&   \revision{\underline{\textbf{37.6}}} &  \revision{\underline{\textbf{31.4}}} &  \revision{\underline{\textbf{24.8}}} \\ \hline\bottomrule
\end{tabular}}}
\begin{tablenotes}
\footnotesize
\item[1] We report the results for Gaussian and SmoothAdv based on pretrained models with the same number of smoothing noise for evaluating \name ($N=10,000$) for a fair comparison. 
\item[2] Boosting is an ensemble method with the base models trained under~\emph{Gaussian},~\emph{SmoothAdv},~\emph{Consistency} and~\emph{MACER}. 
\item[3] The authors use a pretrained BEiT large model~\cite{bao2021beit} in the original paper, and we reimplement DDS on ResNet-50 here and report the results.
\item[4] We use the same smoothed models (i.e., Gaussian and SmoothAdv) used in~\name  for DDS and report the best results.
\end{tablenotes}
\end{threeparttable}
\label{tab:imagenet}
\vspace{-6mm}
\end{table*}


First, for each given (adversarial) input $x$, we will add standard Gaussian smoothing noise to get a set of $x_{rs}$ for certification purposes following~\cite{carlini2022certified}.
We then denoise each noisy input $x_{rs}$ with a diffusion model to get a purified sample $\hat{x}$. We only run the reverse diffusion step once with the diffusion model and directly output the optimal estimate $\hat{x}$ for prediction accuracy and efficiency purposes. In specific, this one-shot reverse diffusion process is implemented to output $x_0 = \frac{1}{\sqrt{\bar{\alpha}_{t}}}(x_{t}-\sqrt{1-\bar{\alpha}_t}\epsilon_{\theta}(x_{t},t))$ given the input $x_t$ and timestep $t$ as shown in~\Cref{alg:denoise}. Usually, the clean instance $x$ is assumed to be in $[0,1]^d$ following prior certification literature; however, the diffusion model expects the input in $[-1,1]^d$, and outputs denoised instance in $[-1,1]^d$. Thus, we start at $x_{t^*}=\sqrt{\bar{\alpha}_{t^*}}(2x_{rs}-1)$ to perform the one-shot reverse step where  $t^*$ is the solution to equation $(2\sigma)^2 = \frac{1-\bar{\alpha}_t}{\bar{\alpha}_t}$.

Second, we perform the \textit{local smoothing} for each purified instance $\hat{x}$; in other words, the local smoothed prediction is provided as $\argmax_{c\in \gY} \sum_{i=1}^{m} F(\textsc{Denoise}(x_{t^*})+\delta_i')/m$ where $\delta_i' \sim \mathcal{N}\left(x, \sigma'^{2} \mathbf{I}\right)$ with $\sigma' \le \sigma$ and $m$ is the number of sampled local smoothing noise. 
As shown in~\Cref{fig:pipeline}, 
the local smoothing noise magnitude $\sigma'$ should be smaller than the smoothing noise level $\sigma$; in the meantime, it
is also related to the robustness of smoothed models to different random noises. For instance, we find that to certify the simple Gaussian augmented model~\cite{cohen2019certified}, we need to set $\sigma'$ to be close to the model smoothing level $\sigma$.
Nevertheless, to certify the advance smoothed model trained with SmoothAdv~\cite{salman2019provably}, setting $\sigma'$ to be around $1/2$ of the $\sigma$ achieves the best certification.
Thus, the local smoothing noise level will also reflect the inherent robustness/stability of the \textit{smoothed} models.


Finally, we will take the majority vote based on these locally smoothed predictions and provide robustness certification for given \textit{smoothed} models following standard randomized smoothing~\cite{cohen2019certified}. 

During the experiment, the first two steps can be wrapped as a single base classifier $f(\cdot)=\textsc{PurifyClassifier}(\cdot;\sigma',m)$ which is shown in~\Cref{alg:purifyf}.
The whole certification process for \name is provided in ~\Cref{alg:certification}, where the function $\textsc{SampleUnderNoise}$ is shown in~\Cref{alg:sample} and the $\textsc{LowerConfBound}(k,n,1-\alpha)$ is a function which returns a  one-sided $(1-\alpha)$ lower confidence bound $\underline{p}$ for the Binomial parameter $p$ given $k \sim \operatorname{Binomial}(n, p)$.

\renewcommand\arraystretch{1.15}
\begin{table*}[h!]
\centering
\caption{\small Certified accuracy of~\emph{Gaussian} with $\sigma=0.50$ under different magnitudes of  local smoothing noise \textit{without} the diffusion-based purification process. The base classifier is trained under noise level $0.50$, and the number of local smoothing noise $m$ is $21$. When $\sigma'$ is set to $``-"$, it represents the standard randomized smoothing setting, indicating that local smoothing is required only when diffusion-based purification is performed.
}
\resizebox{0.6\linewidth}{!}{
\begin{tabular}{ccccccccccc}
\hline
\multirow{2}{*}{Dataset} & \multirow{2}{*}{$\sigma'$} & \multirow{2}{*}{ACR} & \multicolumn{8}{c}{Certified Accuracy under $\ell_2$ Radius $r$} \\
& & & 0.00   & 0.25   & 0.50   & 0.75  & 1.00  & 1.25  & 1.50  & 1.75  \\ \toprule\hline
\multicolumn{1}{c|}{\multirow{4}{*}{CIFAR-10}} & \multicolumn{1}{c|}{-}  &0.534 & \underbf{65.0} & \underbf{54.4} & 41.4 & 32.0 & 23.0 & 15.2 & 9.4 & 5.4\\
\multicolumn{1}{c|}{}                          & \multicolumn{1}{c|}{0.12}  &\underbf{0.538} & \underbf{65.0} & 54.2 & \underbf{42.8} & 33.0 & 23.4 & 15.6 & 10.0 & 5.0 \\
\multicolumn{1}{c|}{}                          & \multicolumn{1}{c|}{0.25}  &0.537 & 64.2 & 54.2 & 41.0 & \underbf{33.2} & \underbf{24.2} & \underbf{16.8} & 10.4 & \underbf{6.2} \\
\multicolumn{1}{c|}{}                          & \multicolumn{1}{c|}{0.50}  &0.433 & 44.8 & 39.0 & 33.8 & 26.0 & 20.4 & 15.0 & \underbf{11.6} & 6.0\\ \hline\hline
\multicolumn{1}{c|}{\multirow{3}{*}{ImageNet}} & \multicolumn{1}{c|}{-}  & \underbf{0.640} & \underbf{56.8} & \underbf{51.2} & \underbf{45.2} & \underbf{41.8} & \underbf{37.0} & \underbf{31.4} & \underbf{24.6} & 0.0  \\
\multicolumn{1}{c|}{}                          & \multicolumn{1}{c|}{0.25}  & 0.333 & 32.0 & 27.4 & 23.8 & 21.6 & 18.6 & 15.0 & 11.8 & 0.0 \\
\multicolumn{1}{c|}{}                          & \multicolumn{1}{c|}{0.50}  & 0.021 & 2.2 & 1.8 & 1.6 & 1.4 & 1.2 & 0.8 & 0.8 & 0.0   \\ \hline \bottomrule
\end{tabular}}
\label{tab:nodiffusion}
\vspace{-5mm}
\end{table*}

\renewcommand\arraystretch{1.13}
\begin{table*}[t]
\centering
\caption{\small Certified accuracy of~\name on CIFAR-10 under different $\ell_2$ radii for smoothed models with noise level $\sigma$ and local smoothing with noise level  $\sigma'$. The base smoothed model selected for each row was pretrained under $\sigma'$. We defer the full certification results w/o diffusion-based purification or local smoothing for smoothed models to~\Cref{adx:compare_baseline} for comparison. 
The number of used local smoothing noise $m$ is $21$, and ACR denotes the average certified radius.}
\resizebox{0.8\linewidth}{!}{
\begin{tabular}{cccccccccccccc}
\toprule\hline
\multirow{2}{*}{Methods} & \multirow{2}{*}{$\sigma$} & \multirow{2}{*}{$\sigma'$} & \multirow{2}{*}{ACR} & \multicolumn{10}{c}{Certified Accuracy (\%) under $\ell_2$ Radius $r$}   \\
 & & & & 0.00 & 0.25 & 0.50 & 0.75 & 1.00 & 1.25 & 1.50 & 1.75 & 2.00 & 2.25 \\ \hline
\multicolumn{1}{c|}{\multirow{9}{*}{\makecell{\name \\ (Gaussian)}}} & \multicolumn{1}{c|}{\multirow{2}{*}{0.25}} & \multicolumn{1}{c|}{0.12}  & 0.543 & \underbf{78.2} & 67.0 & 58.0 & 44.8 & 0.0 & 0.0 & 0.0 & 0.0 & 0.0 & 0.0 \\
\multicolumn{1}{c|}{} & \multicolumn{1}{c|}{} & \multicolumn{1}{c|}{0.25}  & \underbf{0.556} & 76.4 & \underbf{67.2} & \underbf{59.2} & \underbf{47.0} & 0.0 & 0.0 & 0.0 & 0.0 & 0.0 & 0.0 \\ \cline{2-14} 
\multicolumn{1}{c|}{} & \multicolumn{1}{c|}{\multirow{3}{*}{0.50}} & \multicolumn{1}{c|}{0.12}  &0.703 & \underbf{70.4}& 61.2 & 52.4 & 41.6 & 33.0 & 25.4 & 19.4 & 12.2 & 0.0 & 0.0 \\
\multicolumn{1}{c|}{} & \multicolumn{1}{c|}{} & \multicolumn{1}{c|}{0.25}  & 0.745 & 69.6 & \underbf{62.8} & \underbf{55.6} & \underbf{45.2} & 36.4 & 28.2 & 21.6 & 14.6 & 0.0 & 0.0 \\
\multicolumn{1}{c|}{} & \multicolumn{1}{c|}{} & \multicolumn{1}{c|}{0.50} & \underbf{0.760} & 67.4 & 61.6 & 53.4 & 43.6 & \underbf{37.4} & \underbf{31.0} & \underbf{25.0} & \underbf{18.2} & 0.0 & 0.0 \\ \cline{2-14} 
\multicolumn{1}{c|}{} & \multicolumn{1}{c|}{\multirow{4}{*}{1.00}} & \multicolumn{1}{c|}{0.12}  &0.581 & 51.0 & 43.4 & 36.2 & 31.8 & 25.4 & 19.6 & 14.6 & 11.4 & 7.6 & 6.0 \\
\multicolumn{1}{c|}{} & \multicolumn{1}{c|}{} & \multicolumn{1}{c|}{0.25}  &0.638 & \underbf{54.0} & \underbf{46.2} & 39.6 & 33.2 & 26.6 & 21.8 & 17.4 & 12.6 & 9.6 & 7.6  \\
\multicolumn{1}{c|}{} & \multicolumn{1}{c|}{} & \multicolumn{1}{c|}{0.50}  &0.699 & 53.0 & 46.0 & \underbf{40.2} & 34.8 & 29.8 & 23.8 & 20.2 & 15.6 & 12.4 & 10.2 \\
\multicolumn{1}{c|}{} & \multicolumn{1}{c|}{} & \multicolumn{1}{c|}{1.00}  &\underbf{0.784} & 47.8 & 44.6 & 39.6 & \underbf{35.8} & \underbf{31.4} & \underbf{27.0} & \underbf{23.4} & \underbf{19.0} & \underbf{16.4} & \underbf{14.2} \\ \hline\hline
\multicolumn{1}{c|}{\multirow{9}{*}{\makecell{\name \\ (SmoothAdv)}}} & \multicolumn{1}{c|}{\multirow{2}{*}{0.25}} & \multicolumn{1}{c|}{0.12}  &\underbf{0.593} & \underbf{82.8} & \underbf{72.0} & \underbf{62.8} & 49.4 & 0.0 & 0.0 & 0.0 & 0.0 & 0.0 & 0.0 \\
\multicolumn{1}{c|}{} & \multicolumn{1}{c|}{} & \multicolumn{1}{c|}{0.25}  &0.572 & 73.4 & 66.2 & 59.6 & \underbf{51.2} & 0.0 & 0.0 & 0.0 & 0.0 & 0.0 & 0.0 \\ \cline{2-14} 
\multicolumn{1}{c|}{} & \multicolumn{1}{c|}{\multirow{3}{*}{0.50}} & \multicolumn{1}{c|}{0.12}  & 0.733 & \underbf{74.2} & \underbf{64.4} & 54.0 & 45.0 & 33.8 & 28.0 & 18.4 & 13.4 & 0.0 & 0.0\\
\multicolumn{1}{c|}{} & \multicolumn{1}{c|}{} & \multicolumn{1}{c|}{0.25}  &\underbf{0.799} & 68.6 & 61.0 & \underbf{55.4} & \underbf{48.2} & \underbf{41.2} & 34.8 & 26.8 & 18.6 & 0.0 & 0.0 \\
\multicolumn{1}{c|}{} & \multicolumn{1}{c|}{} & \multicolumn{1}{c|}{0.50} & 0.726 & 49.8 & 47.2 & 43.4 & 40.4 & 38.2 & \underbf{36.2} & \underbf{32.0} & \underbf{27.0} & 0.0 & 0.0\\ \cline{2-14} 
\multicolumn{1}{c|}{} & \multicolumn{1}{c|}{\multirow{4}{*}{1.00}} & \multicolumn{1}{c|}{0.12}  &0.584 & 52.0 & 45.6 & 37.6 & 30.4 & 25.2 & 18.8 & 13.6 & 10.2 & 7.6 & 6.0\\
\multicolumn{1}{c|}{} & \multicolumn{1}{c|}{} & \multicolumn{1}{c|}{0.25}  & 0.724 & \underbf{52.6} & \underbf{47.2} & \underbf{40.6}& 36.6 & 31.4 & 26.4 & 21.4 & 15.8 & 12.2 & 10.4  \\
\multicolumn{1}{c|}{} & \multicolumn{1}{c|}{} & \multicolumn{1}{c|}{0.50}  & 0.880 & 45.0 & 42.4 & 39.6 & \underbf{37.4} & \underbf{34.4} & 30.2 & 26.8 & 23.2 & 19.6 & 17.8 \\
\multicolumn{1}{c|}{} & \multicolumn{1}{c|}{} & \multicolumn{1}{c|}{1.00}  & \underbf{0.910} & 44.4 & 41.8 & 39.0 & 36.0 & 33.8 & \underbf{31.0} & \underbf{28.0} & \underbf{25.6} & \underbf{22.0} & \underbf{19.0} \\ \hline\hline
\multicolumn{1}{c|}{\multirow{9}{*}{\makecell{\name \\ (SmoothAdv w/ \\extra data)}}} & \multicolumn{1}{c|}{\multirow{2}{*}{0.25}} & 
\multicolumn{1}{c|}{0.12}  &\underbf{0.624} & \underbf{85.4} & \underbf{76.2} & \underbf{65.6} & 52.0 & 0.0 & 0.0 & 0.0 & 0.0 & 0.0 & 0.0 \\
\multicolumn{1}{c|}{} & \multicolumn{1}{c|}{} & \multicolumn{1}{c|}{0.25}  &0.622 & 79.2 & 72.8 & 65.4 & \underbf{57.0} & 0.0 & 0.0 & 0.0 & 0.0 & 0.0 & 0.0 \\ \cline{2-14} 
\multicolumn{1}{c|}{} & \multicolumn{1}{c|}{\multirow{3}{*}{0.50}} & \multicolumn{1}{c|}{0.12}  &0.771 & \underbf{74.8} & 64.8 & 57.0 & 47.8 & 36.4 & 29.6 & 21.8 & 14.2 & 0.0 & 0.0\\
\multicolumn{1}{c|}{} & \multicolumn{1}{c|}{} & \multicolumn{1}{c|}{0.25}  &\underbf{0.830} & 72.2 & \underbf{65.0} & \underbf{58.8} & \underbf{50.4} & \underbf{43.6} & 33.6 & 26.2 & 18.2 & 0.0 & 0.0 \\
\multicolumn{1}{c|}{} & \multicolumn{1}{c|}{} & \multicolumn{1}{c|}{0.50} &0.794 & 60.8 & 56.0 & 49.6 & 45.6 & 41.4 & \underbf{37.2} & \underbf{31.4} & \underbf{25.2} & 0.0 & 0.0 \\ \cline{2-14} 
\multicolumn{1}{c|}{} & \multicolumn{1}{c|}{\multirow{4}{*}{1.00}} & \multicolumn{1}{c|}{0.12}  &0.623 & 53.8 & 46.6 & 39.0 & 33.0 & 26.2 & 21.4 & 16.4 & 10.8 & 8.2 & 6.4 \\
\multicolumn{1}{c|}{} & \multicolumn{1}{c|}{} & \multicolumn{1}{c|}{0.25}  &0.709 & \underbf{54.0} & \underbf{48.2} & \underbf{43.6} & 37.0 & 30.0 & 24.6 & 20.4 & 15.0 & 11.4 & 8.2 \\
\multicolumn{1}{c|}{} & \multicolumn{1}{c|}{} & \multicolumn{1}{c|}{0.50}  &0.827 & 49.6 & 46.0 & 42.6 & \underbf{39.2} & \underbf{33.8} & 30.0 & 25.6 & 20.6 & 16.8 & 14.2 \\
\multicolumn{1}{c|}{} & \multicolumn{1}{c|}{} & \multicolumn{1}{c|}{1.00}  & \underbf{0.914} & 43.0 & 41.2 & 38.6 & 35.2 & 32.0 & \underbf{30.6} & \underbf{27.2} & \underbf{24.0} & \underbf{21.6} & \underbf{20.0} \\ \hline\hline
\end{tabular}}
\label{tab:cifar_all}
\vspace{-5mm}
\end{table*}

\renewcommand\arraystretch{1.15}
\begin{table*}[ht]
\centering
\caption{
\small Certified accuracy of~\name on ImageNet under different $\ell_2$ radii for 
smoothed models with noise level $\sigma$ and local smoothing with noise level  $\sigma'$. The base smoothed model selected for each row was pretrained under $\sigma'$. We defer the full certification results w/o diffusion-based purification or local smoothing for smoothed models to~\Cref{adx:compare_baseline} for comparison. 
The number of used local smoothing noise $m$ is $21$, and ACR denotes the average certified radius.
}
\resizebox{0.78\linewidth}{!}{
\begin{tabular}{cccccccccccc}
\toprule\hline
\multirow{2}{*}{Architecture} & \multirow{2}{*}{Methods} & \multirow{2}{*}{$\sigma$} & \multirow{2}{*}{$\sigma'$} & \multirow{2}{*}{ACR} & \multicolumn{7}{c}{Certified Accuracy (\%) under $\ell_2$ Radius $r$} \\
& & & & & 0.00    & 0.50    & 1.00    & 1.50    & 2.00   & 2.50   & 3.00   \\ \hline
\multicolumn{1}{c|}{\multirow{12}{*}{ResNet-50}} & \multicolumn{1}{c|}{\multirow{6}{*}{\makecell{\name \\ (Gaussian)}}} & \multicolumn{1}{c|}{\multirow{1}{*}{0.25}} & \multicolumn{1}{c|}{0.25}    & \underbf{0.467} & \underbf{66.2} & \underbf{57.8} & 0.0 & 0.0 & 0.0 & 0.0 & 0.0 \\\cline{3-12} 
\multicolumn{1}{c|}{}&\multicolumn{1}{c|}{} & \multicolumn{1}{c|}{\multirow{2}{*}{0.50}} & \multicolumn{1}{c|}{0.25}  &0.710 & \underbf{57.2} & 50.4 & 41.4 & 31.4 & 0.0 & 0.0 & 0.0 \\
\multicolumn{1}{c|}{}&\multicolumn{1}{c|}{} & \multicolumn{1}{c|}{} & \multicolumn{1}{c|}{0.50}  & \underbf{0.741} & 55.8 & \underbf{50.8} & \underbf{44.2} & \underbf{36.8} & 0.0 & 0.0 & 0.0 \\ \cline{3-12} 
\multicolumn{1}{c|}{}&\multicolumn{1}{c|}{} & \multicolumn{1}{c|}{\multirow{3}{*}{1.00}}  & \multicolumn{1}{c|}{0.25} &0.809 & 43.4 & 38.4 & 31.8 & 24.6 & 19.8 & 16.6 & 12.4\\
\multicolumn{1}{c|}{}&\multicolumn{1}{c|}{} & \multicolumn{1}{c|}{} & \multicolumn{1}{c|}{0.50}  &1.008 & \underbf{48.4} & \underbf{42.4} & 36.2 & 31.6 & 27.2 & 24.0 & 18.4 \\
\multicolumn{1}{c|}{}&\multicolumn{1}{c|}{} & \multicolumn{1}{c|}{} & \multicolumn{1}{c|}{1.00} & \underbf{1.013} & 43.6 & 40.4 & \underbf{37.0} & \underbf{32.8} & \underbf{28.6} & \underbf{25.0} & \underbf{19.8} \\  \cline{2-12} 
\multicolumn{1}{c|}{} & \multicolumn{1}{c|}{\multirow{6}{*}{\makecell{\name \\ (SmoothAdv)}}} & \multicolumn{1}{c|}{\multirow{1}{*}{0.25}} & \multicolumn{1}{c|}{0.25}  &\underbf{0.478} & \underbf{66.2} & \underbf{59.2} & 0.0 & 0.0 & 0.0 & 0.0 & 0.0\\ \cline{3-12} 
\multicolumn{1}{c|}{}&\multicolumn{1}{c|}{} & \multicolumn{1}{c|}{\multirow{2}{*}{0.50}}  & \multicolumn{1}{c|}{0.25}  &\underbf{0.792} & \underbf{59.0} & \underbf{53.4} & \underbf{48.2} & \underbf{39.6} & 0.0 & 0.0 & 0.0 \\
\multicolumn{1}{c|}{}&\multicolumn{1}{c|}{} & \multicolumn{1}{c|}{} & \multicolumn{1}{c|}{0.50}  &0.741 & 54.0 & 49.6 & 44.2 & 38.2 & 0.0 & 0.0 & 0.0\\ \cline{3-12} 
\multicolumn{1}{c|}{}&\multicolumn{1}{c|}{} & \multicolumn{1}{c|}{\multirow{3}{*}{1.00}} & \multicolumn{1}{c|}{0.25} &1.000 & \underbf{48.4} & 43.0 & 36.0 & 31.8 & 27.6 & 23.2 & 16.0\\
\multicolumn{1}{c|}{}&\multicolumn{1}{c|}{} & \multicolumn{1}{c|}{} & \multicolumn{1}{c|}{0.50}  &\underbf{1.087} & 47.6 & \underbf{43.8} & \underbf{39.8} & \underbf{35.0} & \underbf{31.0} & \underbf{25.4} & \underbf{22.4}\\
\multicolumn{1}{c|}{}&\multicolumn{1}{c|}{} & \multicolumn{1}{c|}{} & \multicolumn{1}{c|}{1.00} &0.937 & 37.8 & 34.8 & 32.8 & 30.4 & 27.0 & 23.8 & 21.0 \\ \hline\hline
\multicolumn{1}{c|}{\multirow{9}{*}{BEiT}} & \multicolumn{1}{c|}{\multirow{9}{*}{\makecell{\name \\ (Gaussian)}}} & \multicolumn{1}{c|}{\multirow{2}{*}{0.25}} & \multicolumn{1}{c|}{0.12} & \underbf{0.623} & \underbf{83.8} & \underbf{77.2} & 0.0 & 0.0 & 0.0 & 0.0 & 0.0\\ 
\multicolumn{1}{c|}{}&\multicolumn{1}{c|}{} & \multicolumn{1}{c|}{} & \multicolumn{1}{c|}{0.25} & 0.618 & 82.0 & 76.6 & 0.0 & 0.0 & 0.0 & 0.0 & 0.0  \\ \cline{3-12} 
\multicolumn{1}{c|}{}&\multicolumn{1}{c|}{} & \multicolumn{1}{c|}{\multirow{3}{*}{0.50}} & \multicolumn{1}{c|}{0.12}  & 1.044 & \underbf{79.2} & \underbf{72.6} & 61.8 & 50.2 & 0.0 & 0.0 & 0.0 \\
\multicolumn{1}{c|}{}&\multicolumn{1}{c|}{} & \multicolumn{1}{c|}{} & \multicolumn{1}{c|}{0.25}  & \underbf{1.061} & \underbf{79.2} & 71.8 & \underbf{63.2} & 52.8 & 0.0 & 0.0 & 0.0 \\
\multicolumn{1}{c|}{}&\multicolumn{1}{c|}{} & \multicolumn{1}{c|}{} & \multicolumn{1}{c|}{0.50}&  1.023 & 73.4 & 67.6 & 62.0 & \underbf{53.0}& 0.0 & 0.0 & 0.0 \\\cline{3-12} 
\multicolumn{1}{c|}{}&\multicolumn{1}{c|}{} & \multicolumn{1}{c|}{\multirow{4}{*}{1.00}} & \multicolumn{1}{c|}{0.12}  &1.216 & \underbf{62.2} & 55.0 & 47.8 & 38.2 & 32.4 & 24.2 & 18.6 \\
\multicolumn{1}{c|}{}&\multicolumn{1}{c|}{} & \multicolumn{1}{c|}{} & \multicolumn{1}{c|}{0.25}  & 1.282 & 62.0 & \underbf{57.4} & 49.0 & 40.6 & 34.0 & 27.6 & 20.0 \\
\multicolumn{1}{c|}{}&\multicolumn{1}{c|}{} & \multicolumn{1}{c|}{} & \multicolumn{1}{c|}{0.50}&  \underbf{1.333} & 61.4 & 55.8 & \underbf{49.2} & \underbf{43.0} & \underbf{37.6} & \underbf{31.4} & 22.6 \\
\multicolumn{1}{c|}{}&\multicolumn{1}{c|}{} & \multicolumn{1}{c|}{} & \multicolumn{1}{c|}{1.00}&  1.214 & 50.8 & 47.4 & 43.2 & 38.8 & 35.4 & 31.0 & \underbf{24.8}\\\hline\bottomrule
\end{tabular}}
\label{tab:imagenet_all}
\vspace{-6mm}
\end{table*}

\vspace{-5mm}
\section{Experiments}
\label{sec:evaluation}
\vspace{-2mm}
In this section, we present the evaluation results for our method~\name. We first show the effectiveness of \name compared with the existing baselines. Then we conduct a set of ablation studies to evaluate the influence of different factors, including (1) the importance of local smoothing and diffusion model purification process, (2) the influence of the magnitude of local smoothing noise, and (3) the influence of the number of noise sampled during local smoothing. Concretely, we show that: (i)~\name consistently outperforms all the other baselines under any $\ell_2$ radius in terms of the certified robustness, and the performance can be further improved with a better model (e.g., ViT models); (ii) the significant performance improvement of our method~\name is attributed to the combination of the diffusion model based purification and local smoothing, which verifies the rationale of our method design. We show that without local smoothing or diffusion-based purification, the certified robustness will drop significantly;
iii) using a small magnitude of the local smoothing noise will benefit the certification under small $\ell_2$ radii, while noises with large magnitude provide higher certified accuracy under large radii; iv) the certified accuracy will also be consistently improved with the increase of the number of sampled local smoothing noise;~\revision{v) even with a similar computation cost required by DDS, our method  still outperforms DDS}. The detailed experimental settings and results are shown below.

\begin{figure*}[t]
	\centering
	\includegraphics[width=0.8\linewidth]{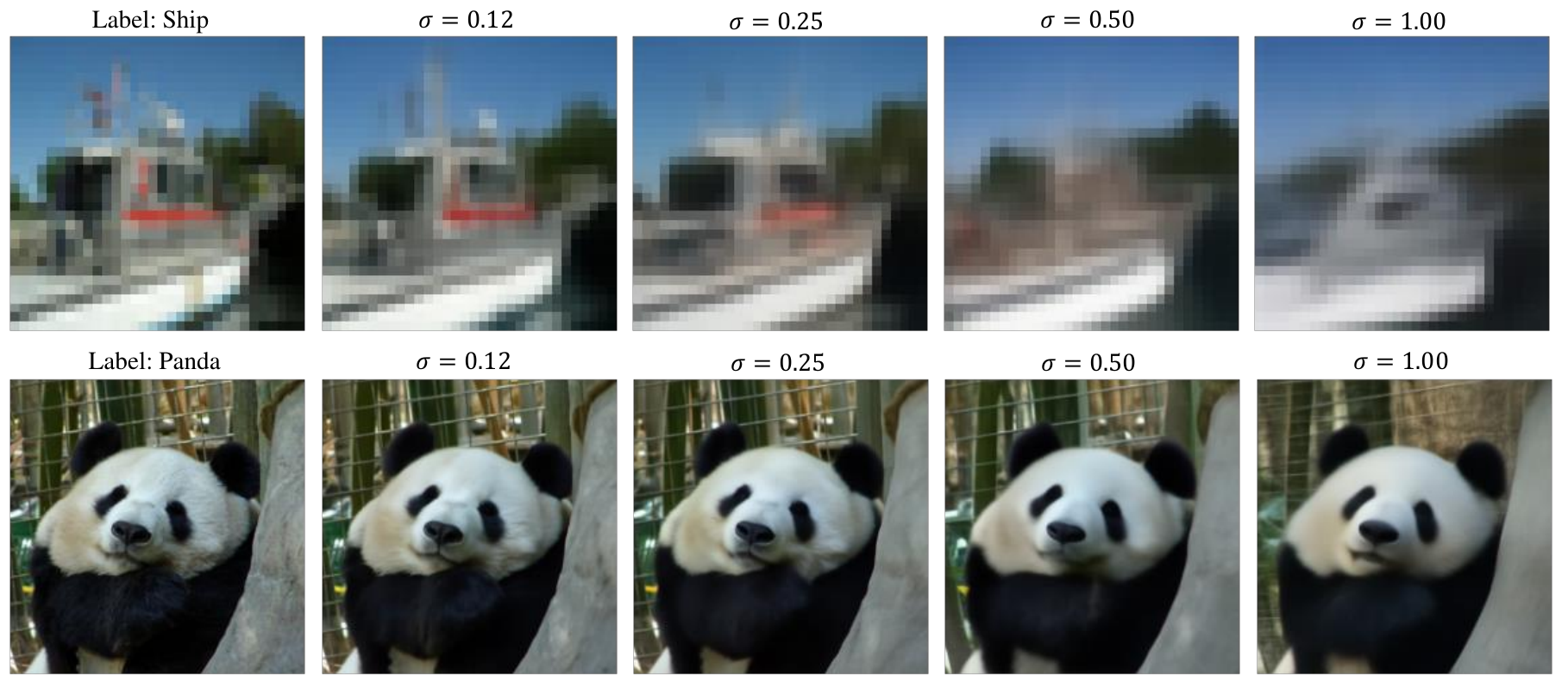}
	\vspace{-5mm}
    \caption{\small Purified images under different magnitudes of Gaussian smoothing noises. Original clean images are shown in the first column. It shows that with higher smoothing noise, the purified image is more blurred.}
    \label{fig:purified_image}
    \vspace{-3mm}
\end{figure*}

\vspace{-7mm}
\subsection{Experimental Setup.}
\label{sec:setup}
\vspace{-1mm}
\paragraph{Dataset and Base Classifiers.}
We conduct the experimental evaluation of our method on datasets CIFAR-10~\cite{krizhevsky2009learning} and ImageNet~\cite{deng2009imagenet}. 
Following the common setting~\cite{cohen2019certified}, we use  ResNet-110 as the base classifier on CIFAR-10 and use ResNet-50~\cite{he2016deep} on ImageNet. To further demonstrate the effectiveness of our method, we also conduct the experiments with a BEiT large model~\cite{bao2021beit} on ImageNet following~\cite{carlini2022certified}. \revision{Specifically, we only finetune the BEiT large model under Gaussian augmentation with $\sigma \in \{0.25,0.50,1.00\}$ using ImageNet-1K based on the self-supervised pretrained model (with intermediate finetuned on ImageNet-22K). For other smoothed models, including ResNet-110 and ResNet-50, we directly use the pretrained ones from~\cite{cohen2019certified,salman2019provably}. } The BEiT large model is fine-tuned with $30$ epochs, and the resolution for both the training and prediction is $224\times224$ instead of $512\times512$ which is used in~\cite{carlini2022certified}; other training hyperparameters of the BEiT are set the same as the finetuning on standard classifier\footnote{\burl{https://github.com/microsoft/unilm/tree/master/beit}}~\revision{and the detailed settings are shown in~\Cref{adx:beit}.}

\vspace{-4mm}
\paragraph{Diffusion Models.} We use the unconditional improved diffusion model\footnote{\burl{https://github.com/openai/improved-diffusion}} from~\cite{nichol2021improved}, which is trained under $L_{hybrid}$ objective, to denoise images from CIFAR-10; and use the unconditional $256\times 256$ guided diffusion model\footnote{\burl{https://github.com/openai/guided-diffusion}} from~\cite{dhariwal2021diffusion} to denoise  images from ImageNet.

\vspace{-6mm}
\paragraph{Baselines.}
We consider eight state-of-the-art $\ell_2$ certifiably robust models as baselines: (1) \emph{Gaussian smoothing}~\cite{cohen2019certified}, which  trains a standard model with Gaussian augmentation for training data; (2) \emph{SmoothAdv}~\cite{salman2019provably}, which  introduces adversarial training into the Gaussian augmented training,~\revision{and it also provides the semi-supervised trained model with extra unlabelled data used in~\cite{carmon2019unlabeled}}; (3) \emph{MACER}~\cite{zhai2019macer}, which tries to maximize the certified radius directly instead of  applying an attack-free algorithm: (4) \emph{Consistency}~\cite{jeong2020consistency}, which adds a consistency regularization term into the training loss; (5) \emph{SmoothMix}~\cite{jeong2021smoothmix}, which combines mixup~\cite{zhang2018mixup} with adversarial training to boost the certified robustness;
(6) \emph{Boosting}~\cite{horvath2021boosting}, which adopts variance-reduced \textit{ensemble} model to generate more consistent prediction; (7)  \emph{Diffusion Denoised Smoothing (DDS(Standard))}~\cite{carlini2022certified}, which leverages diffusion models to remove the added Gaussian smoothing noise and then applies off-the-shelf standard classifiers to predict the purified instances; and (8) \emph{DDS(Smoothed)}, which replaces the standard classifier in DDS(standard) with smoothed classifiers including Gaussian augmented classifier~\cite{cohen2019certified}  and  SmoothAdv~\cite{salman2019provably} respectively, and then selects the maximal certified accuracy among these two.  


\vspace{-5mm}
\paragraph{Certification Details.}
For both CIFAR-10 and ImageNet, we certify a subset of $500$ samples from their test set with confidence $99.9\%$. Besides, each data point is certified with  $N=100,000$ samples of smoothing noise on CIFAR-10 following prior work~\cite{cohen2019certified}, and we use $N=10,000$ on ImageNet following~\cite{carlini2022certified}, the $\sigma$ is selected in $\{0.25,0.50,1.00\}$ for all models.
For each $\sigma$, we try different $\sigma'\le \sigma$ to explore the influence of the magnitude of local smoothing noise. The details are in the~\Cref{adx:localnoise}.

We evaluate our method on the smoothed model trained with two different methods,~\emph{Gaussian} and~\emph{SmoothAdv}, denoted \namens(Gaussian) and \namens(SmoothAdv), respectively. In specific, for the experiments on CIFAR-10 with ResNet-110 and the experiments on ImageNet with ResNet-50, we directly use the pretrained smoothed models from~\emph{Gaussian}~\cite{cohen2019certified}~\footnote{\burl{https://github.com/locuslab/smoothing}} and~\emph{SmoothAdv}~\cite{salman2019provably}~\footnote{\burl{https://github.com/Hadisalman/smoothing-adversarial}}, and the detailed information of the selected pretrained models are deferred to~\Cref{adx:model}. 
For the experiments on ImageNet with BEiT large model, we finetuned the BEiT large model that is pretrained on ImageNet-22K with the method~\emph{Gaussian}.

\vspace{-1mm}
Note that our method does not need to be further finetuned on these purified images. 
All the smoothed models are trained with the clean images $x$, while the prediction during the certification is conducted on the purified images $\hat{x}$. Nevertheless, as shown in~\Cref{fig:purified_image}, the purified images are usually blurrier and less fine-grained than the clean images, and thus the classification accuracy on the purified images will drop a bit when compared to the clean accuracy. So~\cite{carlini2022certified} propose to further finetune the pretrained smoothed models with the purified images. However, such finetuning is quite expensive for ImageNet due to the high cost of the purification. As a compromise, we find that using the local smoothing noise with a magnitude slightly larger than the nominal magnitude can effectively offset the influence of such a distribution shift. 
Empirically, on CIFAR-10, we in fact, add the local smoothing noise with the magnitude of $(\sigma'+0.03)$ under all smoothing level $\sigma$; and we add the local smoothing noise with magnitude $(\sigma'+0.01)$ when $\sigma =0.25$ while adding local smoothing noise with magnitude $(\sigma'+0.02)$ for other $\sigma$ on ImageNet. For reference, we also provide the experiment results without this strategy in~\Cref{adx:shift}.

\vspace{-2mm}
\paragraph{Evaluation Metrics.}
We report the certified accuracy under different $\ell_2$ radius $r$ following the standard certification setting~\cite{cohen2019certified}. In addition, we report the average certified radius (ACR) of  $500$ certified images following~\cite{zhai2019macer}.

\begin{figure*}[h!]
	\centering
	\includegraphics[width=0.8\textwidth]{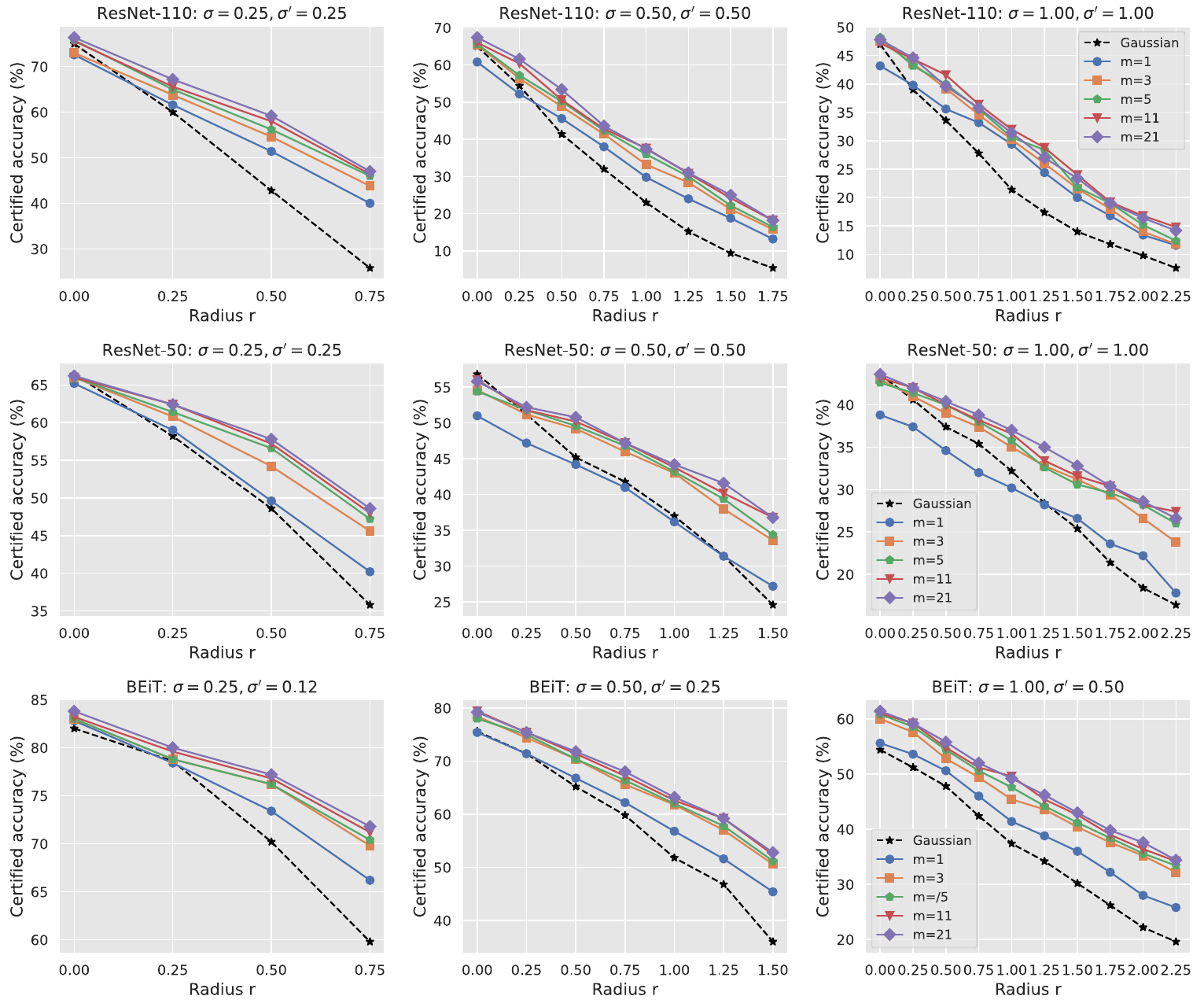}
	\vspace{-5mm}
    \caption{\small The certified accuracy of~\namens(Gaussian) with different numbers of local smoothing noise on CIFAR-10 and ImageNet. We show the results on CIFAR-10 in the first row, while the results on ImageNet with ResNet-50 and BEiT are shown in the second and third rows, respectively.}
    \label{fig:diffm}
    \vspace{-7mm}
\end{figure*}

\renewcommand\arraystretch{1.15}
\begin{table*}[t]
\centering
\begin{algocolor}
\caption{\small Certified accuracy of ResNet-110 on CIFAR-10 under different $\ell_2$ radii with the number of predictions as $100, 000$.}
\label{tab:cifar_sameN}
\resizebox{0.86\linewidth}{!}{
\begin{tabular}{ccccccccccc}
\toprule\hline
\multirow{2}{*}{Method}& \multirow{2}{*}{Setting} & \multicolumn{9}{c}{Certified Accuracy (\%) under $\ell_2$ Radius $r$}   \\
& & 0.00 & 0.25 & 0.50 & 0.75 & 1.00 & 1.25 & 1.50 & 1.75 & 2.00 \\ \hline
\multicolumn{1}{c|}{DDS(Standard)}  & \multicolumn{1}{c|}{$N=100, 000$} & 79.0 & 62.0 & 45.8 & 32.6 & 25.0 & 17.6 & 11.0 & 6.2 & 4.2 \\
\multicolumn{1}{c|}{DDS(Smoothed)}  & \multicolumn{1}{c|}{$N=100, 000$} & 79.8 & 69.9 & 55.0 & 47.6 & 37.4 & 32.4 & 28.6 & 24.8 & 15.4\\ \hline
\multicolumn{1}{c|}{\multirow{3}{*}{\namens(Gaussian)}}  & \multicolumn{1}{c|}{$N=20,000, m=5$} & 77.2 &67.4 & 55.6& 44.4& 35.0& 29.4& 21.8& 18.4& 15.0 \\
 \multicolumn{1}{c|}{} &\multicolumn{1}{c|}{$N=10,000, m=10$} & 76.8 & 66.0 & 56.6 & 42.8 & 36.0 & 29.0 & 23.6 & 18.2 & 16.6 \\
 \multicolumn{1}{c|}{} &\multicolumn{1}{c|}{$N=5,000, m=20$} & 77.2 & 66.8 & 58.2 & 43.8 & 36.6 & 29.4 & 22.0 & 18.0 & 15.0 \\ \hline
\multicolumn{1}{c|}{\multirow{3}{*}{\namens(SmoothAdv)}}  & \multicolumn{1}{c|}{$N=20,000, m=5$} & 82.2 & 71.6 & 62.8 & 49.2 & 39.8 & 35.2 & 29.8 & 24.0 & 22.4 \\
 \multicolumn{1}{c|}{} &\multicolumn{1}{c|}{$N=10,000, m=10$} & 82.8 & 71.0 & 62.4 & 48.4 & 40.0 & 35.4 & 29.6 & 24.6 & 21.0 \\
 \multicolumn{1}{c|}{} &\multicolumn{1}{c|}{$N=5,000, m=20$} & 82.6 & 71.8 & 61.8 & 47.4 & 40.4 & 34.4 & 27.2  & 24.2 &  20.6\\ \hline
 \multicolumn{1}{c|}{\multirow{3}{*}{\makecell{\namens(SmoothAdv)\\ with extra data}}}  & \multicolumn{1}{c|}{$N=20,000, m=5$} & \textbf{86.0} & 75.8 & \textbf{65.6} & \textbf{54.0} & \textbf{41.8} & 35.6 & \textbf{30.2} & 23.8 & \textbf{22.2} \\
 \multicolumn{1}{c|}{} &\multicolumn{1}{c|}{$N=10,000, m=10$} & 85.2 & \textbf{76.0} & 64.2 & 53.8 & \textbf{41.8} & \textbf{36.0} & 28.8 & \textbf{24.4} & 21.4 \\
 \multicolumn{1}{c|}{} &\multicolumn{1}{c|}{$N=5,000, m=20$} & 85.2 & \textbf{76.0} & 64.8  & 49.2  & 41.4  & 34.4 & 26.6 & 23.0 & 20.6 \\ \hline \bottomrule
\end{tabular}
}
\end{algocolor}
\vspace{-5mm}
\end{table*}

\renewcommand\arraystretch{1.15}
\begin{table*}[t]
\centering
\begin{algocolor}
\caption{\small Certified accuracy on ImageNet under different $\ell_2$ radii with the number of predictions as $10,000$.}
\label{tab:imagenet_sameN}
\resizebox{0.82\linewidth}{!}{
\begin{tabular}{ccccccccc}
\toprule\hline
\multirow{2}{*}{Architecture} & \multirow{2}{*}{Method}& \multirow{2}{*}{Setting} & \multicolumn{6}{c}{Certified Accuracy (\%) under $\ell_2$ Radius $r$}   \\
& & & 0.00 & 0.50 & 1.00 & 1.50 & 2.00 & 2.50 \\ \hline
\multicolumn{1}{c|}{\multirow{8}{*}{ResNet-50}} & \multicolumn{1}{c|}{DDS(Standard)}  & \multicolumn{1}{c|}{$N=10,000$} & \textbf{67.4} &49.0 & 33.0 & 22.2 &17.4 &12.8 \\
\multicolumn{1}{c|}{}  & \multicolumn{1}{c|}{DDS(Smoothed)}  & \multicolumn{1}{c|}{$N=10,000$} & 48.0 & 40.6 & 29.6 & 23.8 & 18.6 & 16.0\\ \cline{2-9}
\multicolumn{1}{c|}{}  & \multicolumn{1}{c|}{\multirow{3}{*}{\namens(Gaussian)}}  & \multicolumn{1}{c|}{$N=2,000, m=5$} & 65.4 & 54.8 & 42.4 & 30.2 & 26.8 & 21.0 \\
  \multicolumn{1}{c|}{} & \multicolumn{1}{c|}{} &\multicolumn{1}{c|}{$N=1,000, m=10$}& 65.8 & 55.2 & 42.4 & 30.6 & 27.6 & - \\
  \multicolumn{1}{c|}{}& \multicolumn{1}{c|}{} &\multicolumn{1}{c|}{$N=500, m=20$} & 65.4 & 53.8 & 41.8 & 30.6 & 25.4 & -\\ \cline{2-9}
 \multicolumn{1}{c|}{}& \multicolumn{1}{c|}{\multirow{3}{*}{\namens(SmoothAdv)}}  & \multicolumn{1}{c|}{$N=20,000, m=5$} & 64.0 & \textbf{57.6} & \textbf{46.4} & \textbf{33.8} & \textbf{28.6} & \textbf{23.4} \\
\multicolumn{1}{c|}{} & \multicolumn{1}{c|}{} &\multicolumn{1}{c|}{$N=1,000, m=10$}& 64.6 & 57.2 & 46.0 & 32.8 & 27.8 & - \\
  \multicolumn{1}{c|}{}& \multicolumn{1}{c|}{} &\multicolumn{1}{c|}{$N=500, m=20$} & 65.0 & 56.4 & 45.2 & 32.4 & 26.6 & -\\ \hline\hline
\multicolumn{1}{c|}{\multirow{5}{*}{BEiT}} &  \multicolumn{1}{c|}{DDS(Standard)}  & \multicolumn{1}{c|}{$N=10,000$} & 82.8 &71.1 & 54.3  &  38.1&   {29.5} &  {-} \\
\multicolumn{1}{c|}{}& \multicolumn{1}{c|}{DDS(Smoothed)}  & \multicolumn{1}{c|}{$N=10,000$} & 76.2 & 60.2  & 43.8 & 31.8  &  {22.0} &  {17.8}\\ \cline{2-9}
  \multicolumn{1}{c|}{}& \multicolumn{1}{c|}{\multirow{3}{*}{\namens(Gaussian)}}  & \multicolumn{1}{c|}{$N=2,000, m=5$} & 83.0 & 75.6 & 60.0 & 40.1 & \textbf{34.9} & \textbf{25.7} \\ 
  \multicolumn{1}{c|}{}& \multicolumn{1}{c|}{} &\multicolumn{1}{c|}{$N=1,000, m=10$} & \textbf{83.2} & \textbf{76.2} & \textbf{60.6} & \textbf{40.3} & 34.3 & - \\ 
  \multicolumn{1}{c|}{}& \multicolumn{1}{c|}{} &\multicolumn{1}{c|}{$N=500, m=20$} & 83.4 & 75.0 & 59.6 & \textbf{40.3} & 31.9 & - \\ \hline \bottomrule
\end{tabular}
}
\end{algocolor}
\vspace{-6mm}
\end{table*}

\vspace{-2mm}
\subsection{Main Results}
\label{sec:result}
\vspace{-2mm}
We compare \name with existing baselines on different model architectures and $\sigma$. 
The results on CIFAR-10 and ImageNet are shown in~\Cref{tab:cifar} and~\Cref{tab:imagenet}, respectively.
As we can see, our method \name on smoothed models (i.e., \namens(Gaussian) and \namens(SmoothAdv)) consistently outperforms all other baselines on both CIFAR-10 and ImageNet. In specific, on   pretrained~\emph{Gaussian} model (\namens(Gaussian)) with CIFAR-10 data, \name{}  improves  the certified accuracy largely from $42.8\%$ to $59.2\%$ under $\ell_2$ radius $0.50$, and from $23.0\%$ to $37.4\%$ under larger $\ell_2$ radius $1.00$. 
On the pretrained~\emph{SmoothAdv} model (\namens(SmoothAdv)), \name{} improves the  certified accuracy compared to \namens(Gaussian), achieving state-of-the-art certified robustness. 
On ImageNet with ResNet-50, under $\ell_2$ radius $1.5$, the certified accuracy of \name{} can be improved from $25.4\%$ to $36.8\%$ on the pretrained~\emph{Gaussian} model (i.e., \namens(Gaussian)), and  from $34.6\%$ to $39.6\%$ on the~\emph{SmoothAdv} model (i.e.,  \namens(SmoothAdv)). The performance is further improved on the~\emph{Gaussian} smoothed  BEiT large model, and the certified accuracy is improved from $70.2\%$ to $77.2\%$ under $\ell_2$ radius $0.50$ and  from $36.0\%$ to $53.0\%$ under radius $1.5$.

Note that the only difference between \namens(SmoothAdv) and DDS(Smoothed) is the local smoothing design (DDS(Smoothed) reports the best results of DDS on smoothed models Gaussian and SmoothAdv). When we compare the results of \namens(SmoothAdv) with DDS(Smooth), we find that the certified accuracy drops from $65.6\%$ (\namens(SmoothAdv)) to $55.0\%$ (DDS(Smoothed))) on CIFAR-10, and drop from $48.2\%$ (\namens(SmoothAdv)) to $29.6\%$ (DDS(Smoothed)) on ImageNet under  $\ell_2$ radius $1.0$. Thus, it verifies the importance of local smoothing design in \name.

 
\vspace{-4mm}
\subsection{Ablation studies}
\label{sec:ablation}

\paragraph{Local smoothing w/o diffusion models.}
To verify the importance of our methodology design (i.e., diffusion model   + local smoothing + smoothed classifier), we remove the diffusion model from our system design by directly applying local smoothing noise $\delta'$ on \textit{Gaussian}~\cite{cohen2019certified} (i.e., local smoothing + smoothed classifier). The results are shown in~\Cref{tab:nodiffusion}. We find that local smoothing can not help to improve the certified accuracy of \cite{cohen2019certified}, and the performance even degrades with large $\sigma'$. The reason is that $\|(x+\delta)-x\| \gg \|\hat{x}-x\|$, then with further local smoothing, the image will only be corrupted more, and thus the classification performance will drop naturally. This result further verifies the rationale of our methodology design. 

\vspace{-4mm}
\paragraph{Magnitude of the local smoothing noise.}
To study the influence of the magnitude of local smoothing noise, we conduct experiments among different $\sigma' \in \{0.12, 0.25, 0.50\}$.
The full results of our method on CIFAR-10 and ImageNet are shown in~\Cref{tab:cifar_all} and~\Cref{tab:imagenet_all}, respectively. We can observe that by carefully selecting the $\sigma'$, the performance will be further improved. In addition, we find that to certify a small radius, smaller $\sigma'$ is preferred; however, for certifying a large radius, the choice of $\sigma'$ depends on the model resilience to random noise, i.e., for Gaussian smoothed models, $\sigma'$ needs to be close to $\sigma$ while for SmoothAdv $\sigma'$ can be chosen to be around $\sigma/2$.

\vspace{-4mm}
\paragraph{Number of noise samples for local smoothing.} To evaluate the influence of the number of noise samples $m$ for local smoothing, we evaluate the certified robustness with $m \in {1,3,5,11,21}$. The certified accuracy under different numbers of smoothing noise samples is shown in~\Cref{fig:diffm}. We observe that the certified accuracy will increase monotonically with the number of smoothing noise samples. In practice,  only using five samples ($m=5$)  is enough to achieve non-trivial robustness certification. 

\vspace{-4mm}
\begin{algocolor}
\paragraph{Computation cost.} Additionally, we calculate the time efficiency for certifying one image with different $m$ on one NVIDIA RTX A6000. Specifically, on CIFAR-10, our certification takes $97$s, $111$s, $125$s, $169$s, $240$s for $m=1,3,5,11,21$, respectively; while on ImageNet with ResNet-50, it takes $564$s, $586$s, $608$s, $672$s, $776$s  for  $m=1,3,5,11,21$, respectively. The corresponding computation cost of the one-shot reverse diffusion step on certifying one image is $90$s on CIFAR-10 and $553$s on ImageNet, respectively.
Based on the similar computation costs of the two methods, we can see that the main bottleneck of the computation cost is the reverse diffusion step instead of the local smoothing. 
\end{algocolor}

\vspace{-4mm}
\paragraph{\revision{Number of local smoothing predictions.}}
\revision{For a fair comparison, we constrain the number of local smoothing predictions to be the same as the number of predictions in DDS. In other words, we will maintain $100, 000$ prediction queries on CIFAR10 and  $10, 000$ prediction queries on ImageNet, and the corresponding results are shown in~\Cref{tab:cifar_sameN} and~\Cref{tab:imagenet_sameN} respectively. As we can see, \name  performs significantly better than DDS even with the same computation cost, and setting $m=5$ is already good enough in practice.}

\vspace{-4mm}
\section{Conclusion}
\vspace{-3mm}
In this work, we aim to leverage diffusion-based purification to provide improved certified robustness for \textit{smoothed} models.
We first provide theoretical analysis to show
that the recovered instances from (adversarial) inputs will
be in the bounded neighborhood of the corresponding original instance with high probability, and the ``one-shot''  DDPM can approximate the original instance under mild conditions. 

Based on our analysis, we propose a certifiably robust pipeline,  \name,  for \textit{smoothed} models. In particular,   \name performs
diffusion-based adversarial purification, followed by a local
smoothing step to provide certified robustness for smoothed models. We conduct extensive experiments on different datasets and show that \name can achieve state-of-the-art certified robustness.

One limitation of our method is that it will take more time for certification; however, the main computation cost during certification actually comes from the diffusion step instead of the local smoothing part. \revision{We show that under the same computation cost with DDS, our method still achieves  higher certified robustness and benign accuracy, which provides interesting and promising directions.} Overall, we hope our study sheds light on developing certifiably robust ML models based on diffusion models and smoothed classifiers. 

\vspace{-6mm}
\section*{Acknowledgments}
\vspace{-3mm}
This work is partially supported by the NSF grant No.1910100, No. 2046726, Defense Advanced Research Projects Agency (DARPA) No. HR00112320012, National Aeronautics and Space Administration (NASA) No. 80NSSC20M0229, C3.ai, and Amazon Research Award.

\bibliographystyle{plain}
\bibliography{bib}
\newpage
\appendix
\section{Theorems And Proofs}\label{theoproof}
\vspace{-2mm}
\textbf{Theorem 1.}\textit{
Given a data distribution $p \in \mathcal{C}^2$ and $\mathbb{E}_{\rvx\sim p} [||\rvx||_2^2]< \infty$. Let $p_t$ be the distribution of $\rvx(t)$ generated by \ref{SDE} and suppose $\nabla_{x} \log p_t(x) \le\frac{1}{2}C, \forall t\in [0,T]$. Let $\gamma(t)$ be the coefficient defined in \ref{SDE} and $\overline{\alpha}_t = e^{-\int_0^t \gamma(s)ds}$. Then given an adversarial sample $x_{rs}= x_0+\delta$, solving \ref{reverse-SDE} starting at time $t^*$ and point $x_{t^*}=\sqrt{\bar{\alpha}_{t^*}}x_{rs}$ until time $0$ will generate a reversed random variable $\hat \rvx_0$ such that with a probability of at least $1-\eta$, we have
\begin{align*}
    ||\hat{\rvx}_0-x_0|| \leq ||x_{rs}-x_0||+\sqrt{e^{2 \tau\left(t^*\right)}-1} C_\eta+\tau\left(t^*\right) C
\end{align*}
where $ \tau(t):= \int_0^t \frac{1}{2}\gamma(s) ds$, $C_{\eta}:= \sqrt{ d+2 \sqrt{d \log \frac{1}{\eta}}+2 \log \frac{1}{\eta}}$, and $d$ is the dimension of $x_0$.
}

\renewcommand\arraystretch{1.15}
\begin{table*}[h!]
\centering
\caption{\small Certified accuracy of~\name on ImageNet under different smoothing levels with different magnitudes of the local smoothing noise $\sigma'$ at various $\ell_2$ radius. $\sigma$ is the smoothing noise magnitude on input. The pretrained base classifier was originally trained under $\sigma'$. ACR is the average certified
radius. The number of used local smoothing noise $m$ is $21$ here, and the magnitude of the local smoothing is not shifted during the experiment.}
\resizebox{0.78\linewidth}{!}{
\begin{tabular}{cccccccccccc}
\toprule\hline
\multirow{2}{*}{Architecture} & \multirow{2}{*}{Methods} & \multirow{2}{*}{$\sigma$} & \multirow{2}{*}{$\sigma'$} & \multirow{2}{*}{ACR} & \multicolumn{7}{c}{Certified Accuracy (\%) under $\ell_2$ Radius $r$} \\
 & & & & & 0.00    & 0.50    & 1.00    & 1.50    & 2.00   & 2.50   & 3.00   \\ \hline
\multicolumn{1}{c|}{\multirow{12}{*}{ResNet-50}} & \multicolumn{1}{c|}{\multirow{6}{*}{\makecell{\name \\ (Gaussian)}}} & \multicolumn{1}{c|}{0.25} & \multicolumn{1}{c|}{0.25}  & 0.467 & 66.0 & 57.2 & 0.0 & 0.0 & 0.0 & 0.0 & 0.0\\ \cline{3-12} 
\multicolumn{1}{c|}{} & \multicolumn{1}{c|}{} & \multicolumn{1}{c|}{\multirow{2}{*}{0.50}} & \multicolumn{1}{c|}{0.25}  &0.723 & \underbf{61.4} & \underbf{51.8} & 40.0 & 30.8 & 0.0 & 0.0 & 0.0 \\
\multicolumn{1}{c|}{} & \multicolumn{1}{c|}{} & \multicolumn{1}{c|}{} & \multicolumn{1}{c|}{0.50}  & \underbf{0.730} & 56.8 & 49.8 & \underbf{42.6} & \underbf{36.2} & 0.0 & 0.0 & 0.0  \\ \cline{3-12} 
\multicolumn{1}{c|}{} & \multicolumn{1}{c|}{} & \multicolumn{1}{c|}{\multirow{3}{*}{1.00}} & \multicolumn{1}{c|}{0.25} &0.827 & 44.0 & 36.8 & 32.0 & 26.0 & 21.6 & 15.8 & 11.6\\
\multicolumn{1}{c|}{} & \multicolumn{1}{c|}{} & \multicolumn{1}{c|}{} & \multicolumn{1}{c|}{0.50}  &0.969 & \underbf{44.6} & \underbf{41.8} & \underbf{36.2} & 30.0 & 25.2 & 22.8 & 18.4 \\
\multicolumn{1}{c|}{} & \multicolumn{1}{c|}{} & \multicolumn{1}{c|}{} & \multicolumn{1}{c|}{1.00} & \underbf{0.989} & 42.4 & 38.4 & 35.6 & \underbf{32.2} & \underbf{29.6} & \underbf{24.4} & \underbf{18.8}\\ \cline{2-12} 
\multicolumn{1}{c|}{} & \multicolumn{1}{c|}{\multirow{6}{*}{\makecell{\name \\ (SmoothAdv)}}} & \multicolumn{1}{c|}{0.25} & \multicolumn{1}{c|}{0.25}  &0.475 & 66.8 & 58.2 & 0.0 & 0.0 & 0.0 & 0.0 & 0.0\\ \cline{3-12} 
\multicolumn{1}{c|}{} & \multicolumn{1}{c|}{} & \multicolumn{1}{c|}{\multirow{2}{*}{0.50}} & \multicolumn{1}{c|}{0.25}  &\underbf{0.765} & \underbf{58.2} & \underbf{52.0} & \underbf{46.0} & 36.6 & 0.0 & 0.0 & 0.0 \\
\multicolumn{1}{c|}{} & \multicolumn{1}{c|}{} & \multicolumn{1}{c|}{} & \multicolumn{1}{c|}{0.50}  &0.723 & 53.4 & 48.8 & 43.4 & \underbf{37.4} & 0.0 & 0.0 & 0.0 \\ \cline{3-12} 
\multicolumn{1}{c|}{} & \multicolumn{1}{c|}{} & \multicolumn{1}{c|}{\multirow{3}{*}{1.00}} & \multicolumn{1}{c|}{0.25} &0.938 & \underbf{46.4} & 41.4 & 34.8 & 29.8 & 25.6 & 21.2 & 15.4\\
\multicolumn{1}{c|}{} & \multicolumn{1}{c|}{} & \multicolumn{1}{c|}{} & \multicolumn{1}{c|}{0.50}  &\underbf{1.053} & \underbf{46.4} & \underbf{42.6} & \underbf{38.8} & \underbf{33.6} & \underbf{28.8} & \underbf{25.4} & \underbf{22.4}\\
\multicolumn{1}{c|}{} & \multicolumn{1}{c|}{} & \multicolumn{1}{c|}{} & \multicolumn{1}{c|}{1.00} &0.931 & 37.0 & 34.8 & 33.2 & 29.8 & 26.8 & 24.6 & 21.0 \\ 
\hline\bottomrule
\end{tabular}}
\label{tab:imagenet_all_noshift}
\vspace{-6mm}
\end{table*}

\begin{proof}
We leverage the proof of \cite[Theorem 3.2]{nie2022DiffPure}. By \ref{reverse-SDE}, we can bound
\begin{align*}
\|\hat{\rvx}(0)-x_0\| =&~\left\|x(t^*)+\hat{\rvx}(0)-x(t^*)-x_0\right\| \\
=&~\left\|x\left(t^*\right)+\int_{t^*}^0-\frac{1}{2} \gamma(t)\left[\rvx(t)+2 \nabla_{\rvx(t)} \log p_t(\rvx(t))\right]\right.\\&~ \vphantom{\left\|x\left(t^*\right)+\int_{t^*}^0-\frac{1}{2} \gamma(t)\left[\rvx(t)+2 \nabla_{\rvx(t)} \log p_t(\rvx(t))\right]\right.}\left.d t+\int_{t^*}^0 \sqrt{\gamma(t)} d \overline{\boldsymbol{w}}-x_0\right\| \\
\le &~\left\|x\left(t^*\right)+\int_{t^*}^0-\frac{1}{2} \gamma(t)\rvx(t) d t+\int_{t^*}^0 \sqrt{\gamma(t)} d \overline{\boldsymbol{w}}\right.\\
&~\left.\vphantom{x\left(t^*\right)+\int_{t^*}^0-\frac{1}{2} \gamma(t)\rvx(t) d t+\int_{t^*}^0 \sqrt{\gamma(t)} d \overline{\boldsymbol{w}}}-x_0\right\|+\left\|\int_{t^*}^0-\gamma(t) \nabla_{\rvx(t)} \log p_t(\rvx(t)) d t\right\|
\end{align*}
where the second equation follows from the integration of \ref{reverse-SDE}, and in the last line we have separated the integration of the linear SDE from non-linear SDE involving $\nabla_{\rvx(t)} \log p_t(\rvx(t))$ by using the triangle inequality.

The above linear SDE is a time-varying Ornstein–Uhlenbeck process with a negative time increment that starts from $t=t^*$ to $t=0$ with the initial value set to $x(t^*)$. Denote by $\rvx'(0)$ its solution, from \cite{sarkka2019applied} we know $\rvx'(0)$ follows a Gaussian distribution, where its mean $\boldsymbol{\mu}(0)$ and covariance matrix $\boldsymbol{\Sigma}(0)$ are the solutions of the following two differential equations, respectively:
\begin{align*}
\frac{d \boldsymbol{\mu}}{d t} &=-\frac{1}{2} \gamma(t) \boldsymbol{\mu} \\
\frac{d \boldsymbol{\Sigma}}{d t} &=-\gamma(t) \boldsymbol{\Sigma}+\gamma(t) \boldsymbol{I}_d
\end{align*}
with the initial conditions $\boldsymbol{\mu}\left(t^*\right)=x\left(t^*\right)$ and $\boldsymbol{\Sigma}\left(t^*\right)=\boldsymbol{0}$. By solving these two differential equations, we have that conditioned on $x\left(t^*\right), \rvx^{\prime}(0) \sim \mathcal{N}\left(e^{\tau\left(t^*\right)} x\left(t^*\right),\left(e^{2 \tau\left(t^*\right)}-1\right) \boldsymbol{I}_d\right)$, where $\tau\left(t^*\right):=\int_0^{t^*} \frac{1}{2} \gamma(s) d s$.

Using the reparameterization trick, let $\boldsymbol{\epsilon}\sim \mathcal{N}(\boldsymbol{0}, \boldsymbol{I}_d)$, we have:
\begin{align*}
\rvx^{\prime}(0)-x_0 &=e^{\tau\left(t^*\right)} x\left(t^*\right)+\sqrt{e^{2 \tau\left(t^*\right)}-1} \boldsymbol{\epsilon}-x_0 \\
&=x_{rs}+\sqrt{e^{2 \tau\left(t^*\right)}-1} \boldsymbol{\epsilon}-x_0
\end{align*}
together with $\nabla_{x} \log p_t(x) \le\frac{1}{2}C, \forall t\in [0,T]$, results in that
\begin{align*}
    ||\hat{\rvx}_0-x_0|| \leq ||\sqrt{\left(e^{ \tau\left(t^*\right)}-1\right)} \boldsymbol{\epsilon}+x_{rs}-x_0||+\tau\left(t^*\right) C
\end{align*}
where $\boldsymbol{\epsilon}\sim \mathcal{N}(\boldsymbol{0}, \boldsymbol{I})$. Since $||\boldsymbol{\epsilon}^2||\sim \chi^2(n)$, by the concentration inequality \cite{boucheron2013concentration}, we have
\begin{align*}
     \operatorname{Pr}\left(||\boldsymbol{\epsilon}|| \geq \sqrt{d+2 \sqrt{d \log \frac{1}{\eta}}+2 \log \frac{1}{\eta}}\right) \leq \eta.
\end{align*}
Thus, with probability at least $1-\eta$, we have \begin{align*}
   ||\hat{\rvx}_0-x_0|| \leq ||x_{t^*}-x_0||+\sqrt{e^{2 \tau\left(t^*\right)}-1} C_\eta+\tau\left(t^*\right) C.
   \end{align*}
\end{proof}

\vspace{-5mm}
\noindent\textbf{Theorem 2.}\textit{
Given a data distribution $p \in \mathcal{C}^2$ and $\mathbb{E}_{\rvx\sim p} [||\rvx||_2^2]< \infty$, given a time $t^*$ and point $x_{t^*}=\sqrt{\bar{\alpha}_{t^*}}x_{rs}$, the one-shot reverse diffusion for DDPM \cref{alg:denoise} will output an $\hat x_0$ such that
\begin{align*}
    \left\|\hat x_0 - \mathbb{E}\left[\hat \rvx_0\mid {\hat {\rvx}_{t^*} =  x_{t^*}}\right]\right \|\le \frac{2 \sigma_{t^*}^2 \alpha_{t^*}\left(1-\bar{\alpha}_{t^*}\right)^{3/2}}{\beta_{t^*}^2 \sqrt{\bar{\alpha}_{t^*}}}\cdot \ell_{t^*}(x_{t^*})
\end{align*}
where $\hat \rvx_0, \hat \rvx_t$ are random variables generated by \ref{reverse-SDE}, $ \mathbb{P}\left(\hat{\rvx}_0 = x| {\hat {\rvx}_t = x_{t^*}}\right) \propto p(x) \cdot  \frac{1}{\sqrt{\left(2\pi\sigma^2_t\right)^n}} \exp\left({\frac{-|| x -x_{t^*}||^2_2}{2\sigma^2_t}}\right)$ and $\sigma_t^2 = \frac{1-\alpha_t}{\alpha_t}$ is the variance of Gaussian noise added at time $t$ in the diffusion process.}
\vspace{-3mm}
\begin{proof}
Given time $t^*$ and point $x_{t^*}=\sqrt{\bar{\alpha}_{t^*}}x_{rs}$ is equivalent to that in the formula of $\ell_{t^*}(x_{t^*})$, given $\sqrt{\bar{\alpha}_{t^*}} \rvx_0+\sqrt{1-\bar{\alpha}_{t^*}}  \boldsymbol{\epsilon} = x_{t^*}$. Then the conditional distribution of $\rvx_0$ will be $ \mathbb{P}\left({\rvx}_0 = x| {{\rvx}_t = x_{t^*}}\right) \propto p(x) \cdot  \frac{1}{\sqrt{\left(2\pi\sigma^2_t\right)^n}} \exp\left({\frac{-|| x -x_{t^*}||^2_2}{2\sigma^2_t}}\right)$. By the assumptions that $p \in \mathcal{C}^2$ and $\mathbb{E}_{\rvx\sim p} [||\rvx||_2^2]< \infty$, we know the diffusion process by \ref{SDE} and the reverse process by \ref{reverse-SDE} follows the same distribution, thus we also have $ \mathbb{P}\left(\hat{\rvx}_0 = x| {\hat {\rvx}_t = x_{t^*}}\right) \propto p(x) \cdot  \frac{1}{\sqrt{\left(2\pi\sigma^2_t\right)^n}} \exp\left({\frac{-|| x -x_{t^*}||^2_2}{2\sigma^2_t}}\right)$. Further note that
\begin{align*}
     \ell_{t^*}(x_{t^*})=&~\mathbb{E}_{\rvx_0, \boldsymbol{\epsilon}}\left[\frac{\beta_{t^*}^2}{2 \sigma_{t^*}^2 \alpha_{t^*}\left(1-\bar{\alpha}_{t^*}\right)}\left\| \boldsymbol{\epsilon}- {\epsilon}_\theta\left(x_{t^*}, {t^*}\right)\right\|^2\right.\\
    &~\qquad \left|\vphantom{\frac{\beta_{t^*}^2}{2 \sigma_{t^*}^2 \alpha_{t^*}\left(1-\bar{\alpha}_{t^*}\right)}\left\| {\epsilon}- {\epsilon}_\theta\left(x_{t^*}, {t^*}\right)\right\|^2}\sqrt{\bar{\alpha}_{t^*}} \rvx_0+\sqrt{1-\bar{\alpha}_{t^*}}  {\epsilon} = x_{t^*}\right]\\
    =&~\mathbb{E}_{\rvx_0, \boldsymbol{\epsilon}}\left[\frac{\beta_{t^*}^2}{2 \sigma_{t^*}^2 \alpha_{t^*}\left(1-\bar{\alpha}_{t^*}\right)}\left\| \boldsymbol{\epsilon}- {\epsilon}_\theta\left(x_{t^*}, {t^*}\right)\right\|^2\right.\\
    &~\qquad \left|\vphantom{\frac{\beta_{t^*}^2}{2 \sigma_{t^*}^2 \alpha_{t^*}\left(1-\bar{\alpha}_{t^*}\right)}\left\| {\epsilon}- {\epsilon}_\theta\left(x_{t^*}, {t^*}\right)\right\|^2}\rvx_{t^*} = x_{t^*}\right]\\
    =&~\mathbb{E}_{\hat \rvx_0, \boldsymbol{\epsilon}}\left[\frac{\beta_{t^*}^2}{2 \sigma_{t^*}^2 \alpha_{t^*}\left(1-\bar{\alpha}_{t^*}\right)}\left\| \boldsymbol{\epsilon}- {\epsilon}_\theta\left(x_{t^*}, {t^*}\right)\right\|^2\right.\\
    &~\qquad \left|\vphantom{\frac{\beta_{t^*}^2}{2 \sigma_{t^*}^2 \alpha_{t^*}\left(1-\bar{\alpha}_{t^*}\right)}\left\| {\epsilon}- {\epsilon}_\theta\left(x_{t^*}, {t^*}\right)\right\|^2}\hat \rvx_{t^*} = x_{t^*}\right].
\end{align*}
Under the condition $\sqrt{\bar{\alpha}_{t^*}} \rvx_0+\sqrt{1-\bar{\alpha}_{t^*}}  \boldsymbol{\epsilon} = x_{t^*}$, which is equivalent to that $\sqrt{\bar{\alpha}_{t^*}} \hat \rvx_0+\sqrt{1-\bar{\alpha}_{t^*}}  \boldsymbol{\epsilon} = x_{t^*}$ where $\boldsymbol{\epsilon} \sim \mathcal{N}(\boldsymbol{0}, \boldsymbol{I})$, there is a one-to-one corresponding between $\hat \rvx_0$ and $\boldsymbol{\epsilon}$, and 
\begin{align*}
    \left\| \boldsymbol{\epsilon}- {\epsilon}_\theta\left(x_{t^*}, {t^*}\right)\right\|^2=&~\left\|\frac{x_{t^*}-\sqrt{\bar{\alpha}_{t^*}} \hat \rvx_0}{\sqrt{1-\bar{\alpha}_{t^*}}}- \frac{x_{t^*}-\sqrt{\bar{\alpha}_{t^*}} {\epsilon}_\theta\left(x_{t^*}, {t^*}\right)}{\sqrt{1-\bar{\alpha}_{t^*}}}\right\|\cdot\\
    =&~ \left\| \hat \rvx_0 - \hat x_0\right\|\cdot \frac{\bar{\alpha}_{t^*}}{\sqrt{1-\bar{\alpha}_{t^*}}}.
\end{align*}
Therefore,
\begin{align*}
     \ell_{t^*}(x_{t^*})=&~\mathbb{E}_{\rvx_0, \boldsymbol{\epsilon}}\left[\frac{\beta_{t^*}^2 s\sqrt{\bar{\alpha}_{t^*}}}{2 \sigma_{t^*}^2 \alpha_{t^*}\left(1-\bar{\alpha}_{t^*}\right)^{3/2}}\left\| \hat \rvx_0 - \hat x_0\right\|\right|\left.\vphantom{\frac{\beta_{t^*}^2 \bar{\alpha}_{t^*}}{2 \sigma_{t^*}^2 \alpha_{t^*}\left(1-\bar{\alpha}_{t^*}\right)^{3/2}}}\rvx_{t^*} = x_{t^*}\right]\\
     \ge &~ \frac{\beta_{t^*}^2 \sqrt{\bar{\alpha}_{t^*}}}{2 \sigma_{t^*}^2 \alpha_{t^*}\left(1-\bar{\alpha}_{t^*}\right)^{3/2}}\cdot  \left\|\hat x_0 - \mathbb{E}\left[\hat \rvx_0\mid {\hat {\rvx}_{t^*} =  x_{t^*}}\right]\right \|,
\end{align*}
where the last inequality is by Jensen's inequality \cite{boyd2004convex}.
\end{proof}

\vspace{-10mm}
\section{Experiment detail}
\label{adx:experiment_detail}

\vspace{-3mm}
\subsection{Pretrained smoothed models}
\label{adx:model}
We directly use the smoothed models trained under~\emph{Gaussian} from~\cite{cohen2019certified}; while for the models trained under~\emph{SmoothAdv}, we select the best-performing models from~\cite{salman2019provably}, and the detailed specification of the hyperparameters for each picked model is shown in~\Cref{tab:selected_smoothadv} and~\Cref{tab:cifar_self}. Notice, when the smoothing level is with $\sigma = 0.25$, the empirical best magnitude of the local smoothing noise for the SmoothAdv model is $\sigma' = \sigma/2 \approx 0.12$. However, there is no pretrained model provided on ImageNet for SmoothAdv under noise level $0.12$; then, as an alternative, we in fact, use the SmoothAdv model trained with smaller $\epsilon$ which is $64$ instead of $512$ when certifying on the smoothing level $\sigma = 0.25$ for ImageNet.

\vspace{-5mm}
\subsection{The setting of the magnitude of the local smoothing noise}
\label{adx:localnoise}
On CIFAR10, we test the $\sigma' \in \{0.12,0.25\}$ for the smoothing level $\sigma = 0.25$, the $\sigma' \in \{0.12,0.25,0.50\}$ for the smoothing level $\sigma = 0.50$, and the $\sigma' \in \{0.12,0.25,0.50,1.00\}$ for the smoothing level $\sigma = 1.00$.

On ImageNet, for ResNet-50, we test the $\sigma' \in \{0.25\}$ for the smoothing level $\sigma = 0.25$, the $\sigma' \in \{0.25,0.50\}$ for the smoothing level $\sigma = 0.50$, and the $\sigma' \in \{0.25,0.50,1.00\}$ for the smoothing level $\sigma = 1.00$. And for BEiT large model, we test the $\sigma' \in \{0.12,0.25\}$ for the smoothing level $\sigma = 0.25$, the $\sigma' \in \{0.12,0.25,0.50\}$ for the smoothing level $\sigma = 0.50$.

\vspace{-4mm}
\begin{algocolor}
\subsection{Finetuning details of BEiT model}
\label{adx:beit}
We finetune the BEiT with the checkpoints that are self-supervised pretrained and then intermediate fine-tuned on ImageNet-22k and train it with Gaussian augmentation with $\sigma\in \{0.25,0.50,1.00\}$ in $30$ epochs. The batch size is $32$, the learning rate is $2e-5$, the update frequency is $2$, the number of warmup epochs is $5$, the layerwise learning rate decay is $0.9$, the drop path is set to $0.4$, and the weight decay is set to $1e-8$.

\end{algocolor}

\vspace{-5mm}
\renewcommand\arraystretch{1.1}
\begin{table}[ht]
\caption{\small Detailed specification of the hyperparameters for the selected SmoothAdv models on CIFAR-10 and ImageNet.}
\centering
\resizebox{0.85\linewidth}{!}{
\begin{tabular}{cccccc}
\toprule\hline
Dataset                                  & $\sigma$ & Method & \# steps & $\epsilon$ & $m$ \\ \hline
\multirow{4}{*}{CIFAR-10} &     0.12     &   PGD     & 10         & 64           & 4    \\
&       0.25   & PGD        &   10       &  255          &  8    \\
&  0.50         &  PGD      &  10        &      512      & 2    \\ 
& 1.00         & PGD        & 10          &     512       & 2     \\ \hline\hline
\multirow{3}{*}{ImageNet}&      0.25    & DNN        & 2          &     512       &   1  \\
&    0.50      &    PGD    &      1    &       255     &    1 \\
&    1.00        &     PGD   &     1     &       512     &     1\\ \hline\bottomrule
\end{tabular}}
\label{tab:selected_smoothadv}
\vspace{-4mm}
\end{table}

\vspace{-5mm}
\renewcommand\arraystretch{1.1}
\begin{table}[ht]
\caption{\small Detailed specification of the hyperparameters for the selected SmoothAdv models with self-training on CIFAR-10.}
\centering
\resizebox{0.85\linewidth}{!}{
\begin{tabular}{cccccc}
\toprule\hline
Dataset                                  & $\sigma$ & Method & \# steps & $\epsilon$ & weight \\ \hline
\multirow{4}{*}{\makecell{CIFAR-10\\(Self-training)}} &     0.12     &   PGD     & 8         & 64           & 1.0    \\
&       0.25   & PGD        &   4       &  127          &  1.0     \\
&  0.50         &  PGD      &  2        &      255      & 0.5    \\ 
& 1.00         & PGD        & 8          &     512       & 0.5     \\ \hline\bottomrule
\end{tabular}}
\label{tab:cifar_self}
\vspace{-5mm}
\end{table}

\vspace{-3mm}
\section{Influence of the shifting on the magnitude of local smoothing noise}
\label{adx:shift}



We provide the experiment results without magnitude shifting for ResNet-50 on ImageNet in~\Cref{tab:imagenet_all_noshift} for comparison.

\renewcommand\arraystretch{1.15}
\begin{table*}[h]
\centering
\caption{Certified accuracy on ImageNet for the smoothed model w/o purification and local smoothing, namely, with the standard randomized smoothing setting, under different smoothing levels $\sigma$.}
\resizebox{0.9\linewidth}{!}{
\begin{tabular}{ccccccccccc}
\toprule\hline
\multirow{2}{*}{Architecture} & \multirow{2}{*}{Methods} & \multirow{2}{*}{$\sigma$}& \multirow{2}{*}{ACR} & \multicolumn{7}{c}{Certified Accuracy (\%) under $\ell_2$ Radius $r$}   \\
 & & & & 0.00 & 0.50 & 1.00 & 1.50 & 2.00 & 2.50 & 3.00 \\ \hline
\multicolumn{1}{c|}{\multirow{6}{*}{ResNet-50}} & \multicolumn{1}{c|}{\multirow{3}{*}{Gaussian}} & 0.25  &  0.413 & \underbf{66.4} & \underbf{48.6} & 0.0 & 0.0 & 0.0 & 0.0 & 0.0 \\ 
 \multicolumn{1}{c|}{} & \multicolumn{1}{c|}{} & 0.50  &  0.640 & 56.8 & 45.2 & \underbf{37.0} & 24.6 & 0.0 & 0.0 & 0.0\\
  \multicolumn{1}{c|}{} & \multicolumn{1}{c|}{} & 1.00 &  \underbf{0.789 }& 43.6 & 37.4 & 32.2 & \underbf{25.4} & \underbf{18.4} & \underbf{13.8} & \underbf{10.4}  \\ 
\cline{2-11}
 \multicolumn{1}{c|}{} &\multicolumn{1}{c|}{\multirow{3}{*}{SmoothAdv}} & 0.25 & 0.446 & \underbf{66.6} & \underbf{52.6} & 0.0 & 0.0 & 0.0 & 0.0 & 0.0 \\ 
 \multicolumn{1}{c|}{} & \multicolumn{1}{c|}{} & 0.50 &  0.707 & 54.6 & 48.6 & \underbf{42.2} & \underbf{34.6} & 0.0 & 0.0 & 0.0 \\ 
  \multicolumn{1}{c|}{} & \multicolumn{1}{c|}{} & 1.00  &  \underbf{0.886} & 38.6 & 35.0 & 32.0 & 28.8 & \underbf{25.2} & \underbf{21.4} & \underbf{18.8}\\ 
 \hline\hline
\multicolumn{1}{c|}{\multirow{2}{*}{BEiT}}& \multicolumn{1}{c|}{\multirow{2}{*}{Gaussian}} & 0.25 & 0.579 & \underbf{82.0} & \underbf{70.2} & 0.0 & 0.0 & 0.0 & 0.0 & 0.0 \\
  \multicolumn{1}{c|}{} & \multicolumn{1}{c|}{} & 0.50  & \underbf{0.910} & 75.6 & 65.2 & \underbf{51.8} & \underbf{36.0} & 0.0 & 0.0 & 0.0 \\
 \hline\hline
\end{tabular}}
\label{tab:imagenet_baseline}
\end{table*}

\renewcommand\arraystretch{1.15}
\begin{table*}[h]
\centering
\caption{Certified accuracy on CIFAR-10 for the selected pretrained smoothed model w/o purification and local smoothing, namely, with the standard randomized smoothing setting, under different smoothing levels $\sigma$.}
\resizebox{0.9\linewidth}{!}{
\begin{tabular}{ccccccccccccc}
\toprule\hline
\multirow{2}{*}{Methods} & \multirow{2}{*}{$\sigma$}& \multirow{2}{*}{ACR} & \multicolumn{10}{c}{Certified Accuracy (\%) under $\ell_2$ Radius $r$}   \\
 & & & 0.00 & 0.25 & 0.50 & 0.75 & 1.00 & 1.25 & 1.50 & 1.75 & 2.00 & 2.25 \\ \hline
\multicolumn{1}{c|}{\multirow{3}{*}{Gaussian}} & 0.25  & 0.423 & \underbf{75.0} & \underbf{60.0} & \underbf{42.8} & 25.8 & 0.0 & 0.0 & 0.0 & 0.0 & 0.0 & 0.0  \\
  \multicolumn{1}{c|}{} & 0.50  &0.534 & 65.0 & 54.4 & 41.4 & \underbf{32.0} & \underbf{23.0} & 15.2 & 9.4 & 5.4 & 0.0 & 0.0  \\
   \multicolumn{1}{c|}{} & 1.00  &\underbf{0.539} & 47.0 & 39.0 & 33.6 & 27.8 & 21.4 & \underbf{17.4} & \underbf{14.0} & \underbf{11.8} & \underbf{9.8} & \underbf{7.6}  \\
 \hline\hline
 \multicolumn{1}{c|}{\multirow{3}{*}{SmoothAdv}} & 0.25  & 0.546 & \underbf{73.6} & \underbf{66.8} & \underbf{57.2} & \underbf{47.2} & 0.0 & 0.0 & 0.0 & 0.0 & 0.0 & 0.0 \\
  \multicolumn{1}{c|}{} & 0.50  & 0.698 & 50.2 & 46.2 & 44.4 & 39.6 & \underbf{37.6} & \underbf{32.8} & \underbf{28.8} & \underbf{23.6} & 0.0 & 0.0   \\
   \multicolumn{1}{c|}{} & 1.00  & \underbf{0.844} & 45.0 & 41.0 & 38.0 & 34.8 & 32.2 & 28.8 & 25.0 & 22.8 & \underbf{19.4} & \underbf{16.8} \\
 \hline\hline
 \multicolumn{1}{c|}{\multirow{3}{*}{\makecell{SmoothAdv\\(w/ extra data)}}} & 0.25  & 0.600 & \underbf{80.8} & \underbf{71.4} & \underbf{63.2} & \underbf{52.6} & 0.0 & 0.0 & 0.0 & 0.0 & 0.0 & 0.0  \\
  \multicolumn{1}{c|}{} & 0.50  &0.753 & 61.8 & 56.8 & 49.0 & 45.2 & \underbf{39.4} & \underbf{32.2} & \underbf{26.2} & 20.0 & 0.0 & 0.0  \\
   \multicolumn{1}{c|}{} & 1.00  &  \underbf{0.868} & 43.6 & 40.6 & 37.6 & 35.0 & 32.6 & 28.8 & 25.8 & \underbf{22.2} & \underbf{20.2} & \underbf{18.4}   \\
 \hline\hline
\end{tabular}}
\label{tab:cifar_baseline}
\end{table*}

\section{The detailed certification results of the baselines for the selected pretrained smoothed model}
\label{adx:compare_baseline}
For better comparison, we also provide the certification results for the selected pretrained smoothed model w/o diffusion model and local smoothing, namely, with the standard smoothing certification~\cite{cohen2019certified}. The result on ImageNet is shown in~\cref{tab:imagenet_baseline} and the result on CIFAR-10 is shown in~\cref{tab:cifar_baseline}.

\end{document}